\documentclass[10pt,journal,compsoc]{IEEEtran}

\ifCLASSOPTIONcompsoc
\usepackage[nocompress]{cite}
\else
\usepackage{cite}
\fi
\usepackage{epsfig}
\usepackage{array}
\usepackage{graphicx}
\usepackage{amsmath,bm}
\usepackage{amssymb}
\usepackage{textcomp}
\usepackage{caption}
\usepackage{subcaption}
\usepackage[table]{xcolor}
\usepackage{makecell,multirow,diagbox}
\usepackage{setspace}
\usepackage{threeparttable}
\usepackage{float}
\usepackage{stfloats}

\usepackage{algorithm}
\usepackage{algorithmicx}
\usepackage{algpseudocode}
\usepackage{booktabs}
\usepackage{arydshln}

\usepackage[pagebackref=true,breaklinks=true,letterpaper=true,colorlinks,bookmarks=false,citecolor = black]{hyperref}
\newcommand{\BLUE}[1]{\textcolor[rgb]{0.0,0.0,1.0}{#1}}
\newcommand{\RED}[1]{\textcolor[rgb]{1.0,0.0,.0}{#1}}

\newcommand{\etal}[1]{\textit{et al.}}
\newcommand{\eg}[1]{\textit{e.g.}}
\newcommand{\etc}[1]{\textit{etc.}}

\begin{document}
	\bstctlcite{IEEEexample:BSTcontrol}
	
	\title{Cyclic Differentiable Architecture Search}
	
	\author{Hongyuan~Yu,
		Houwen~Peng,
		Yan~Huang,
		Jianlong~Fu,
		Hao~Du,
		Liang~Wang, %~\IEEEmembership{Fellow,~IEEE}
		Haibin Ling
		
		\IEEEcompsocitemizethanks{
			\IEEEcompsocthanksitem Hongyuan Yu, Yan Huang and Liang Wang are with the Center for Research on Intelligent Perception and Computing (CRIPAC), National Laboratory of Pattern Recognition (NLPR), the Center for Excellence in Brain Science and Intelligence Technology (CEBSIT) and the University of Chinese Academy of Sciences (UCAS), Beijing, China.\protect\\
			Email: hongyuan.yu@nlpr.ia.ac.cn, \{yhuang, wangliang\}@nlpr.ia.ac.cn
			
			\IEEEcompsocthanksitem Houwen Peng and Jianlong Fu are with the Microsoft Research. %, MSRA, China.\protect\\
			% E-mail:  \{houwen.peng,jianf\}@microsoft.com, lzuqer@gmail.com
			Email: \{houwen.peng,jianf\}@microsoft.com
			\IEEEcompsocthanksitem Hao Du is with the Department of Computer Science, City University of Hong Kong, Hong Kong, China. Email: haodu8-c@my.cityu.edu.hk 
			\IEEEcompsocthanksitem Haibin Ling is with the Department of Computer Science, Stony Brook University. Email: haibin.ling@stonybrook.edu  \protect 
			\IEEEcompsocthanksitem Work performed when Hongyuan Yu and Hao Du were interns of Microsoft Research. 
			Houwen Peng and Liang Wang are the corresponding authors.
			\protect \\
		}% <-this % stops an unwanted space
	}


	\markboth{Journal of \LaTeX\ Class Files,~Vol.~*, No.~*, *~*}%
	{Shell \MakeLowercase{\textit{et al.}}: CDARTS}

	%\IEEEpubid{\makebox[\columnwidth]{\hfill 0000--0000/00/\$00.00~\copyright~2014 IEEE}%


	\IEEEtitleabstractindextext{%
		\begin{abstract}
			% In this paper, we study Neural Architecture Search (NAS) for image classification, which allows the automatic design of deep neural networks. 
			Differentiable ARchiTecture Search, \emph{i.e.}, DARTS, has drawn great attention in neural architecture search. It tries to find the optimal architecture in a shallow search network and then measures its performance in a deep evaluation network.
			%Recent differentiable Neural Architecture Search (NAS) methods try to find the optimal architecture in a shallow search network, and then separately measure its performance in a deep evaluation network. 
			The independent optimization of the search and evaluation networks, however, leaves a room for potential improvement by allowing interaction between the two networks.
			% To address this issue, we propose a novel Cyclic Differentiable ARchiTecture Search framework, dubbed CDARTS. Considering the structure difference, CDARTS builds a cyclic feedback mechanism between the search and evaluation networks with introspective distillation. 
			To address the problematic optimization issue, we propose new joint optimization objectives and a novel Cyclic Differentiable ARchiTecture Search framework, dubbed CDARTS. Considering the structure difference, CDARTS builds a cyclic feedback mechanism between the search and evaluation networks with introspective distillation.
			%This leads to the optimization of architecture search is independent of the target evaluation network, and the discovered architecture is sub-optimal. 
			%To address this issue, we present a novel differentiable architecture search framework with introspective distillation (CDARTS), which enables the search process of deep neural networks in a unified framework. 
			First, the search network generates an initial architecture for
			evaluation, and the weights of the evaluation network are optimized. Second, the architecture weights in the search network are further optimized by the label supervision in classification, as well as the regularization from the evaluation network through feature distillation. Repeating the above cycle results in a joint optimization of the search and evaluation networks and thus enables the evolution of the architecture to fit the final evaluation network.
			The experiments and analysis on CIFAR, ImageNet and NATS-Bench demonstrate the effectiveness of the proposed approach over the state-of-the-art ones.
			Specifically, in the DARTS search space, we achieve 97.52\% top-1 accuracy on CIFAR10 and 76.3\% top-1 accuracy on ImageNet. In the chain-structured search space, we achieve 78.2\% top-1 accuracy on ImageNet, which is 1.1\% higher than EfficientNet-B0.
			% Moreover, the experiments on NAS-Bench-201, chain-structured search space show the generality and robustness of the proposed method. 
			%We provide all information (training and test code, documentations) needed to reproduce the proposed approach and the results at 
			%Our code and models are available at 	{\textcolor{blue}{\small{\url{https://github.com/researchmm/CDARTS}}}}.
			Our code and models are publicly available at \href{https://github.com/microsoft/Cream}{https://github.com/microsoft/Cream}.
		\end{abstract}
		
		\begin{IEEEkeywords}
			Cyclic, Introspective Distillation, Differentiable Architecture Search, Unified Framework
		\end{IEEEkeywords}
	}

	\maketitle

	% is not selected <OR> if conference mode is selected - because compsoc
	% conference papers position the abstract like regular (non-compsoc)
	\IEEEdisplaynontitleabstractindextext

	% \fi
	\IEEEpeerreviewmaketitle
	
	\IEEEraisesectionheading{\section{Introduction}\label{sec:introduction}}
	\IEEEPARstart{D}{eep} learning has enabled remarkable progress in a variety of vision tasks over the past years. One crucial factor for this progress is the design of novel neural network architectures. Most of current employed architectures are designed by human experts, which is time-consuming and error-prone. Because of this, there is a growing interest in automatic Neural Architecture Search (NAS) for vision tasks, such as image recognition~\cite{NASNet,chang2020data}, object detection~\cite{ghiasi2019fpn,chen2019detnas} and semantic segmentation~\cite{liu2019auto,nekrasov2019fast}.
	
	Differentiable architecture search, \emph{i.e.}, DARTS~\cite{DARTS}, has become one of the most popular NAS pipelines nowadays due to its relatively low computational cost and competitive performance	~\cite{ren2021comprehensive}. Different from previous methods~\cite{NASNet,NAS_RL,PNAS,kandasamy2018neural} that search over a discrete set of candidate architectures, DARTS relaxes the search space to be continuous, so that the architecture can be optimized by the gradient descent. The efficiency of gradient-based optimization reduces the search cost from thousands of GPU days to a few. According to the recent NAS survey~\cite{Survey,Survey2,ren2021comprehensive}, due to the simplicity and elegance of the DARTS architecture, the research work related to DARTS is quite rich~\cite{cai2018proxylessnas,PDARTS,yang2021towards}. Moreover, gradient optimization in a continuous search strategy is an important trend of NAS.
	%The efficient gradient-based optimization allows DARTS to reduce the search time from thousands of GPU days to a few.
	
\begin{figure}[t]
		\vspace{-0.2cm}
		\centering
		\includegraphics[scale=1.3]{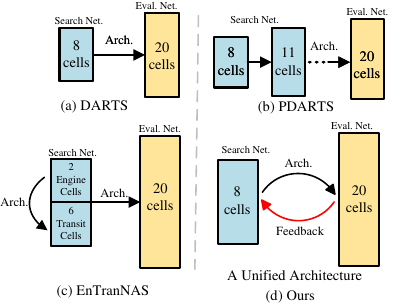}
		%\vspace{-0.2cm}
		\caption{Comparisons of prior DARTS and our CDARTS. In prior DARTS~\cite{DARTS}, PDARTS~\cite{PDARTS}, and EnTranNAS~\cite{yang2021towards}, the target evaluation network does not engage in the architecture search. In contrast, our CDARTS combines the search and evaluation networks into a joint optimization framework. The \textcolor[rgb]{0.67,0.847,0.92}{\bf blue} and \textcolor[rgb]{1,0.8,0.2}{\bf gold}  boxes indicate the search and evaluation networks, respectively. %(Best viewed in color).
		}
		\label{fig:intro}
		\vspace{-0.4cm} 
	\end{figure}

	% However, limited by the high GPU memory consumption,
	Existing DARTS approaches have to divide the search process into two steps: search and evaluation, as shown in Fig.~\ref{fig:intro} (a). The search step employs a small network to discover the optimal cell\footnote{Cell is a basic building block for network construction. It consists of convolution, pooling, nonlinearity and a prudent selection of connections.} structures, and the evaluation step stacks the discovered cells to construct a large network for final evaluations.
	This results in the optimization of search process is independent from the target evaluation network. 
    As shown in Fig.~\ref{fig:intro} (b), PDARTS~\cite{PDARTS} tried to alleviate the gap by gradually deepening the search network. In Fig.~\ref{fig:intro}(c), EnTranNAS~\cite{yang2021towards} built the search network by combining the evaluation network module and the search network module to reduce the gap.
	Besides, SNAS~\cite{xie2018snas} and GDAS~\cite{GDAS} applied Gumbel-softmax and modified straight-through Gumbel-Softmax to relieve the gap caused by discretization. Furthermore, AutoHAS~\cite{AutoHAS} augmented GDAS with an entropy term to search for both hyperparameters and architectures.	
	However, those methods still separates the search and evaluation process.
	As a consequence, the performance of discovered architectures in the search network has limited correlation with the actual performance of the evaluation network. On the other hand, there are several recent works casting doubt on the effectiveness of DARTS. Li and Talwalkar~\cite{randomNAS} observe that even a simple random search method can find architectures outperforming the original DARTS. Zela \emph{et al.}~\cite{Understanding} and Liang \emph{et al.}~\cite{liang2019darts+} show  that DARTS is prone to degenerate to networks filled with parametric-free operations, \emph{e.g.} skip connections, leading to a poor performance of the searched architecture. Wang \emph{et al.}~\cite{wang2021rethinking} find that the problematic optimization of DARTS results in the learned architecture weights are insufficient to reflect the relative ranking of different architectures during evaluation~\cite{RandomSPOS,hard,yang2021towards,xie2020weight}.
	
	%\textcolor{red}{Wang \emph{et al.}~\cite{wang2021rethinking} find that the operation associated with larger architecture parameters $\alpha$ does not necessarily result in higher validation accuracy after discretization.} They further show mathematically that the domination of skip connection observed in DARTS is in fact a reasonable outcome of the search network’s optimization. Thus, the problematic optimization of DARTS results in the learned architecture weights are insufficient to reflect the relative ranking of different architectures during evaluation~\cite{RandomSPOS,hard,yang2021towards,xie2020weight}}. 
	
	% To alleviate these issues, we propose a differentiable architecture search method with introspective distillation, dubbed CDARTS. 
	To alleviate these issues, we propose a \textit{cyclic differentiable architecture search} method, dubbed CDARTS.  CDARTS integrates the search and evaluation networks into a unified architecture, and jointly trains the two networks in a cyclic way. 
	As visualized in Fig.~\ref{fig:intro}(c), the shallow search network provides the best intermediate architecture to the evaluation network. In turn, the search network gets the feedback from the evaluation network (with higher model capacity due to more layers) by introspective distillation. The distillation is introspective because it does not require introducing third-party models, such as human-designed architectures, to serve as the teacher models. This is crucial for practical applications because there is usually no available teacher model in new tasks. As proved by Liu \etal~\cite{liu2019search} and Li \etal~\cite{blockwise}, the knowledge of a network model not only lies in the network parameters but also the network structure. The introspective distillation could transfer the knowledge embedded in parameters as well as the the knowledge inside the structure from the evaluation network to the search network. 
	
	% When the teacher model, the evaluation network, is gradually updated along with the training process, which allows the evolution of cell structures to fit the final evaluation network.  
	% the evaluation network (with higher model capacity due to more layers) feedback the feature knowledge by introspective distillation into the search network to enhance the search of architectures. 
	
	% It should be noted that the knowledge of a network model not only lies in the network parameters but also the network structure~\cite{liu2019search,blockwise}. Thus, the feature distillation could transfer the parameters as well as the structure knowledge of the evaluation network to the search network.
	
	% Note that we call it introspective distillation because it does not require introducing third-party models, 
	
	% It brings the ability to perceive the quality of the discovered architecture to the search network. 
	
	Moreover, instead of training the search and evaluation networks separately, we propose a joint learning algorithm to optimize the integrated architecture in a cyclic manner. It consists of a separate learning and a joint learning stage. The separate learning stage aims to optimize the search and evaluation networks to have good initializations. The joint learning stage is to update the architecture and network weights alternatively. Specifically, the joint learning stage first optimizes the architecture weights and the evaluation network weights by jointly training the search and evaluation networks. Then, it optimizes the search network weights according to the updated architecture weights. These two learning stages are performed alternatively, 
	leading to a cyclic optimization between the search and evaluation networks. 
	Eventually, the target evaluation network obtains a shaped architecture tailored by the search network. 
	%\BLUE{The proposed jointly training method make it possible to directly optimize the evaluation network, which relives the problematic optimization~\cite{wang2021rethinking} of DARTS.}
	The proposed jointly training scheme allows the direct optimization of the evaluation network, reliving the problematic optimization of DARTS~\cite{wang2021rethinking}.
	
	We evaluate our CDARTS algorithm on image classification task and conduct experiments on CIFAR~\cite{krizhevsky2009learning}, ImageNet~\cite{russakovsky2015imagenet}, as well as the recently proposed NATS-Bench benchmark~\cite{dong2021nats}.
	The experiments demonstrate that, on CIFAR, CDARTS achieves superior performance to existing state-of-the-art DARTS approaches. 
	%such as PDARTS~\cite{PDARTS}, PCDARTS~\cite{PCDARTS} and FairDARTS~\cite{chu2019fairdarts}. 
	On the large-scale ImageNet, the proposed CDARTS also shows its superiority in the DARTS family, while achieving comparable performance to MobileNet-V3~\cite{mobilenetv3}, which blends automatic search techniques with human intuition. Moreover, for a fair comparison with other NAS algorithms, such as one-shot~\cite{SETN,GDAS} and reinforcement learning-based~\cite{REINFORCE,ENAS} 
	approaches, we conduct an experiment on the NATS-Bench benchmark, which provides a fixed search space with a unified training setting. The experiment shows that CDARTS achieves competitive performance compared with 10 recent prevailing NAS approaches.
	% The analysis on NAS-Bench-201 demonstrates the efficacy of our approach is due to the improved search stability. We observe that the performance of discovered architectures is improved gradually along with the increase of search epochs, and eventually reaches stable and converged. 
	% The correlation between the learned architecture hyperparameters and the true actual performance of discovered architectures is enhanced as well.
	Last but not the least, we also apply our introspective distillation search method to the chain-structured search space~\cite{Survey} and achieve a gain of 1.6\% over EfficientNet-B0~\cite{tan2019efficientnet} on ImageNet with a similar model size. 
	We further transfer the discovered architectures to the downstream object detection task and semantic segmentation task. % The new searched network surpasses MobileNet-V3~\cite{mobilenetv3} by 5.9\% on COCO~\cite{coco} detection dataset.
	For object detection, we get an AP of 36.2 on the COCO validation set, which surpasses the state-of-the-art MobileNetV3 by 6.3\%. And for semantic segmentation, we obtain an mIoU of 78.2\% on the cityscapes validation set and 40.4\% on ADE20K validation set, which are 1.9\% and 3\% superior than the recent SegFormer~\cite{xie2021segformer}, respectively.

	\vspace{-0.05cm}
	Our main contributions are summarized as below:
	\vspace{-0.15cm}
	%The main contributions of this work is  as follows.
	\begin{itemize}
		%\item We integrate the search and evaluation networks into a cyclic learning framework, and thus address the separated optimization issue and improve the search stability. Accordingly, we propose a joint learning algorithm to optimize the integrated architecture in a cooperative manner.
		\item We rethink the bi-level optimization for NAS and empirically show that the performance of discovered architectures in the search network has a limited correlation with the actual performance of the evaluation network, leading to the instability issue. We propose brand-new joint optimization objectives for differentiable one-shot NAS.
		\item  We integrate the search and evaluation networks into a unified framework, and propose a cyclic learning algorithm. Such algorithm addresses the problematic optimization issue in previous differentiable methods~\cite{DARTS,PDARTS,chu2019fairdarts}, thus being able to improve the stability of the search algorithm.
		\item We propose an introspective distillation mechanism to transfer the knowledge embedded in both parameters and structure from the evaluation network to the search network. In contrast to classical distillation methods~\cite{hinton2015distilling,blockwise}, such introspective method does not require a predefined third-party teacher model, which is more flexible in practice. %in practical scenarios.
		%\item Extensive experimental results on three architecture search space and vision tasks demonstrate that the proposed method not only improve the search generality and robustness, but also significantly improves the performance on different benchmarks.
		\item Extensive experiments on three types of search spaces verify the robustness of the proposed method. Moreover, the results of the searched architectures in the object detection and semantic segmentation tasks demonstrate their generality and transferability.
	\end{itemize}

	\section{Related Work}
	\label{Related}
	Deep neural networks (DNN) have played an important role in computer vision tasks such as image recognition~\cite{russakovsky2015imagenet,shen2015multi,yang2020dynamical}, object detection~\cite{liu2020deep,ren2015faster}, segmentation~\cite{liu2010robust,yang2020sognet,li2019expectation}, tracking~\cite{tracking_pami}, and so on~\cite{lin_pami,li2021towards,zhong2018joint}. However, designing effective and efficient neural architectures is a labor-intensive process that may require lots of trials by human experts. Hence, automatic neural architecture search (NAS)~\cite{han2015learning,adanet,smithson2016neural,saxena2016convolutional} has emerged in recent years. In this section, we first summarize the search space, and then discuss the search strategies in recent NAS methods. At last, we briefly review some distillation methods which are most related to our method.
	
	%\vspace{-0.3cm}
	\subsection{Search Space}
	The search space defines all possible architectures in principle. A well designed search space can simplify the search. Current search spaces mainly fall into two categories: chain-structured space and cell-based space.
	%Similar to Wistuba et al.~\cite{Survey2}, we divided the search space into chain-structured search space and cell-base search space.
	
	In the chain-structured search space, the neural architecture can be represented by a sequence of stacked layers. Baker \emph{et al.}~\cite{baker2016designing} consider a set of layers that includes convolutions, pooling and dense connections. The corresponding hyperparameter settings contain the number of filters, kernel size, stride and pooling size. Zoph \emph{et al.}~\cite{NAS_RL} define a relaxed version of the chain-structured search space by introducing skip connections between blocks. %in the chain-structured architecture. 
	Xie \emph{et al.}~\cite{xie2017genetic} extend the search space by replacing feature concatenation operations with summations when merging features. %Most recently, the development of human designed lightweight networks have aroused researchers' attention~\cite{howard2017mobilenets,sandler2018mobilenetv2,zhang2018shufflenet,shufflenetv2}. Among them, inverted residual blocks are used to search mobile platform neural architectures~\cite{tan2019mnasnet}. 
	Most recently, the inverted residual blocks are introduced to facilitate the search of neural architectures for mobile platforms. 
	% MobileNets~\cite{howard2017mobilenets,sandler2018mobilenetv2} employ depth-wise convolution as basic components and propose inverted residual block to achieve a better trade-off between performance and model size. ShuffleNets~\cite{zhang2018shufflenet,shufflenetv2} further introduce several guidelines for designing efficient networks. 
	% Take the advantage of human priori of the chain-structured search space greatly boost the development of NAS. For example, 
	%MNAS~\cite{tan2019mnasnet} use reinforcement learning to search mobile platform neural architectures under the MobileNetV2~\cite{sandler2018mobilenetv2} structure. ProxylessNAS~\cite{cai2018proxylessnas} and other related works~\cite{spos,blockwise} adapt MobileNetV2 as their baseline structure in the one-shot search framework. 
	MNAS~\cite{tan2019mnasnet}, ProxylessNAS~\cite{cai2018proxylessnas} and other related methods~\cite{blockwise,bignas,dai2020fbnetv3} searches architectures over the space consisting of the lightweight inverted bottleneck MBConvs~\cite{sandler2018mobilenetv2}. %Furthermore, the state-of-the-art methods~\cite{ofa,moga} adapts MobileNetV3~\cite{mobilenetv3} architecture to build a new search space and achieve impressive results under mobile settings.
	MobileNetV3~\cite{mobilenetv3} and OFA~\cite{ofa} extend the space with the Swish activation function and squeeze-excitation modules, thus achieving impressive results under mobile settings.
	%perform architecture search over the search space consisting of mobile inverted bottleneck MBConv [11] and squeeze-excitation modules
	%The performance of the baseline search space and the discovered architecture show a strong correlation. 
	%Recent state-of-the-art works~\cite{ofa,moga} have adopted MobileNetV3~\cite{mobilenetv3} architecture to build their search space and achieve impressive results under mobile settings. 
	% cell-based search space
	
	Although models searched based on chain-structured search space have shown strong representation ability, it generally consumes extensive computation cost to exhaustively cover the search space during training. Motivated by hand-crafted architectures consisting of repeated motifs~\cite{he2016deep,huang2017densely,hu2018senet}, Zoph et al ~\cite{NASNet} and Zhong et al~\cite{zhong2018practical} firstly build networks by repeating the searched cells which is easy to scale the model size by adjusting the number of cell~\cite{real2019regularized}.
	
	% A well-designed cell module enables transferability between datasets~\cite{ENAS,DARTS}. It is also easy to scale down or up the model size by adjusting the number of cell repeats~\cite{real2019regularized}.
	
	% in the chain-structured search space, it is difficult to directly search large-scale architectures, which usually have much stronger representation ability. Motivated by hand-crafted architectures consisting of repeated motifs~\cite{he2016deep,huang2017densely,hu2018senet}, Zoph et al ~\cite{NASNet} and Zhong et al~\cite{zhong2018practical} firstly build networks by repeating the searched cells. Different from the architectures found in the chain-structured search space, cells can be stacked to build big models~\cite{real2019regularized}. ENAS~\cite{ENAS} and DARTS~\cite{DARTS} further carefully design the cell-based search space and achieve impressive results. }
	
	The development of NAS algorithms in different kinds of search spaces indicates that a better search space is essential for designing search strategies. Recently, some works~\cite{hard,Understanding} reveal that even random sampled cells can get pretty good results under those carefully designed search space. Therefore, several NAS  benchmarks~\cite{nasbench101,nasbench1shot1,dong2021nats} are released to eliminate the bias of search space and relieve the evaluation burden. In this paper, we apply our method to both chain-structured search space and cell-base search space and achieve promising results on CIFAR~\cite{krizhevsky2009learning} and ImageNet~\cite{russakovsky2015imagenet}. Furthermore, we verify the efficiency of our method on NATS-Bench benchmark~\cite{dong2021nats}.

	\subsection{Search Strategy}
	Early NAS approaches focus on searching a network motif by using either reinforcement learning~\cite{NASNet,NAS_RL} or evolution algorithms~\cite{real2019regularized,xie2017genetic,wang2018evolving,liang2018evolutionary,sun2018evolving,dong2018dpp} and build a complete network for evaluation by stacking the motif. Unfortunately, these approaches require to train hundreds or thousands of candidate architectures from scratch, resulting in barely affordable computational overhead.
	%\emph{e.g.}, hundreds or even thousands of GPU days. 
	Thus, several follow-up works are proposed to speed-up training by reducing the search space~\cite{PNAS,before_pnas}. %or inheriting weights~\cite{}.
	
	More recent works resort to the weight-sharing \emph{one-shot} model to amortize the searching cost~\cite{randomNAS,ENAS,SMASH,saxena2016convolutional}. The key idea of one-shot approach is to train a single over-parameterized model, and then share the weights between sampled child models. This weight sharing mechanism allows the searching of a high-quality architecture within a few GPU days~\cite{bender2018understanding,wu2019fbnet}. Single path one-shot model families~\cite{ofa,randomNAS,spos,moga,blockwise} propose to train a single sampled path of the one-shot model in each iteration, rather than the entire over-parameterized model. Once the training process is finished, the child models can be ranked by the shared weights. However, \cite{Understanding} finds that the weight sharing strategy results in inaccurate performance evaluation of the candidate architecture, making it difficult for the one-shot NAS to identify the best architecture. FairNAS~\cite{fairnas} and PC-NAS~\cite{li2020improving} also demonstrate that neural architecture candidates  based on these parameter sharing methods also cannot be adequately trained, which leads to an inaccurate ranking of architecture candidates . There are few works leveraging knowledge distillation~\cite{hinton2015distilling} to boost the training of the super network, and they commonly introduce additional large models as teachers. More specifically, OFA~\cite{ofa} pretrains the largest model in the search space and use it to guide the training of other subnetworks, while  DNA~\cite{blockwise} directly employs the third-party EfficientNet-B7~\cite{tan2019efficientnet} as the teacher model. These search algorithms will become infeasible if there is no available pretrained model, especially when the search task and data are entirely new. In contrast to those methods, our introspective distillation method does not require a predefined third-party teacher model, which is more flexible in practice.
	
	Differentiable architecture search, \emph{i.e.}, DARTS~\cite{DARTS}, is another representative one-shot model. 
	Instead of searching over a discrete set of candidate architectures, DARTS relaxes the search space to be continuous, so that the architecture can be optimized by the efficient gradient descent. 
	Despite DARTS has achieved promising performance by only using orders of magnitude less computation resources, some issues remain. ProxylessNAS~\cite{cai2018proxylessnas} argues that the optimization objectives of search and evaluation networks are inconsistent in DARTS~\cite{DARTS}. Thus it employs a binary connection to tackle the issue. PDARTS~\cite{PDARTS} points out the depth gap problem of DARTS, and thereby presents a progressive learning approach, which gradually increases the number of layers in the search network. Moreover, PCDARTS~\cite{PCDARTS} addresses the problem of high GPU memory cost through introducing a partially-connected strategy in network optimization.
	SNAS~\cite{xie2018snas} and GDAS~\cite{GDAS} tried to relieve the discretization gap with modified Gumbel-Softmax. Furthermore, AutoHAS~\cite{AutoHAS} augmented GDAS with an entropy term to search for both hyperparameters and architectures. ISTA-NAS~\cite{yang2020ista} formulated neural architecture search as a sparse coding problem to avoid the post-processing of DARTS. Moreover, EnTranNAS~\cite{yang2021towards}  proposed an architecture derivation method to replace the hand-crafted post-processing rules.
	
	Our proposed approach differs from existing DARTS approaches~\cite{DARTS,PDARTS,PCDARTS,chu2019fairdarts} in two major ways. One is that our approach integrates the search and evaluation networks into a unified architecture. Also, we build up connections to transfer the parameters as well as the structure knowledge of the evaluation network to the search network.
	% allowing the information exchange during the search of architectures. 
	The other difference is the searching algorithm. In contrast to previous methods where the optimization of architecture weights is independent from the evaluation network, our approach enables the evaluation network to guide the search of architectures by joint training.

	\subsection{Knowledge Transfer Between Architectures}
	Transferring knowledge between architectures during the search is widely used in NAS. A simple way is sharing weights across architectures. Such way can largely reduce training cost because the knowledge from previous searches is reused. TAPAS~\cite{istrate2019tapas} starts with a simple architecture and scales up it based on a prediction model. %T-NAML~\cite{wong2018transfer} uses a  network pre-trained on ImageNet and makes various decisions to adapt it to target datasets. The reinforcement learning method is used to optimize neural architectures across several data sets simultaneously.
	T-NAML~\cite{wong2018transfer} adapts a pre-trained network to target datasets  with reinforcement learning, which makes decisions on optimizing neural architectures across datasets simultaneously. 
	XferNet~\cite{wistuba2019xfernas} accelerates NAS by transferring architecture knowledge across different tasks. NATS~\cite{peng2019efficient} introduces a practical neural architecture transformation search algorithm for object detection. It explores architecture space based on  existing networks and reusing their weights without searching and re-training. Instead of only inheriting weights from previously-trained networks,  FNA++~\cite{fang2020fna++} adapt both the architecture and parameters of a seed network, which makes it possible to use NAS for segmentation and detection more efficiently. NAT~\cite{lu2021neural} is also designed to efficiently generate task-specific models under multiple searching objectives. %It learns task-specific supernets from which specialized subnets can be sampled without any additional training by iteratively adapting a pre-trained supernet and searching for task-specific subnets simultaneously.}
	
	%However, the scope of weight inheritance is relatively narrow.
	Knowledge distillation~\cite{hinton2015distilling} is another widely used technique for knowledge transfer between architectures. It compresses the ``dark knowledge'' of a well trained larger model to a smaller one \cite{Cream}. An enormous amount of approaches have been proposed to fortify the efficiency of student models’ learning capability.
	% Fitnets~\cite{Fitnets} firstly propose the concept of hint learning, aiming at reducing the distance between feature maps of students and teachers. Agoruyko et al.~\cite{zagoruyko2016paying} try to align the features of attention regions from the point of view of attention mechanism. Knowledge distillation also has potential in the other domains. Wu et al.~\cite{wu2020learning} focus on open-set problems via massive knowledge distillation. 
	% Zhang et al. attempt to train a lightweight network for image super-resolution via the importance learning of individual pixels. Song et al.~\cite{song2020simultaneous} propose an adaptation strategy to address the dehazing problem of camera-captured images by distilling the cross-domain knowledge. Liu et al.~\cite{liu2019search} discovers the knowledge of a given neural network depends on both the parameters and the architecture. Thus, they develop a architecture-aware knowledge distillation approach that finds student models that are best for distilling the given teacher model. 
	Recently, one-shot methods try to improve the search process by using the supervision of other high-capacity networks, and they commonly introduce additional large models as teachers. BIGNAS~\cite{bignas} and OFA~\cite{ofa} pretrain the largest model in the search space and use it to guide the training of other subnetworks, while DNA~\cite{blockwise} directly employs the third-party EfficientNet-B7~\cite{tan2019efficientnet} as the teacher model. These search algorithms will become infeasible if there is no available pretrained model, especially when the search task and data are entirely new. Different from these methods, the teacher network of our method is generated by the model itself. Hence, our model does not need to spend extra time to train a teacher network alone and thereby is more flexible.
	
	\section{Methodology}
	\vspace{-0.cm}
	\label{Method}
	In this section, we present the Cyclic Differentiable Architecture Search method, \emph{i.e.}, CDARTS. We first briefly review the standard gradient-based differentiable method, which trains the search and evaluation networks independently~\cite{DARTS,PDARTS,PCDARTS,chu2019fairdarts}. Then, we propose the cyclic search method, which learns the search and evaluation networks jointly. At the end, we present the implementation details of the network architecture in our method.
	\vspace{-0.cm}
	\subsection{Preliminary: Differentiable Architecture Search}
	\vspace{-0.cm}
	\label{sec:preliminary}
	
	The goal of Differentiable Architecture Search, \emph{i.e.}, DARTS, ~\cite{DARTS} is to search a cell motif, which can be stacked repeatedly to construct a convolutional network. A cell is a directed acyclic graph consisting of an ordered sequence of $N$ nodes $\{x_i\}_{i=0}^{N-1}$. Each node $x_i$ is a latent representation (\emph{e.g.}, a feature map), while each directed edge $(i,j)$ is associated with one operation $o^{(i,j)}$ which transforms information from $x_i$ to $x_j$. Let $\mathcal{O}$ denote the operation space, consisting of a set of candidate operations, such as convolution, max-pooling, skip-connection, etc. Each operation represents a function $o(\cdot)$ to be performed on $x_i$.
	To make the search space continuous, DARTS relaxes the choice of a particular operation to a softmax over all possible operations~\cite{DARTS}:
	\vspace{-0.cm}
	\begin{equation} 
	\centering
	\bar{o}^{(i,j)} (x_i) = \sum_{o \in \mathcal{O}} \frac{\exp (\alpha_o^{(i,j)})}{\sum_{o' \in \mathcal{O}} \exp (\alpha_{o'}^{(i,j)})} \cdot o(x_i),
	\label{Eq_darts_node}
	\vspace{0.1cm}
	\end{equation}
	where the operation weights for a pair of nodes $(i, j)$ are parameterized by the architecture weights $\alpha(i,j)$ of
	dimension $|\mathcal{O}|$. Here, the parameter $\alpha$ is the encoding of architectures to be optimized.
	An intermediate node of the cell is computed based on all of its predecessors as $x_j=\sum_{i<j} \bar{o}^{(i,j)} (x_i)$, and the output node $x_{N-1}$ is obtained by concatenating all the intermediate nodes in the channel dimension. 
	% (excluding the input nodes)
	
	With the definition of a cell, the search of the optimal architecture becomes to solve the following bilevel optimization problem:
	\vspace{-0.cm}
	\begin{equation}
	\centering
	\begin{aligned}
	&\min_\alpha\quad  \mathcal{L}_{val}  (w_S^*, \alpha),
	\\
	\mathrm{s.t.}\quad & w_{S}^*  = \mathop{\arg \min}_{w_S} \mathcal{L}_{train} (w_S, \alpha),  
	\end{aligned}
	\label{Eq_darts_s}
	\vspace{0.1cm}
	\end{equation}
	where $\alpha$ contains the architecture weights optimized on the validation data (\emph{val}), and $w_S$ denotes the parameters of the \emph{search} network learnt on the training data (\emph{train}). Eq.(\ref{Eq_darts_s}) amounts to optimize the network and architecture weights,~\emph{i.e.}, $w_S$ and $\alpha$, in an alternative way. 
	After getting the optimal architecture, DARTS constructs a new evaluation network by stacking the discovered neural cells and retrains from scratch over the training of the target task.
	
	Without loss of generality, here we only consider one cell from a simplified search space consists of two operations: (skip, conv). According to Wang et al.~\cite{wang2021rethinking}, the current estimation of the optimal feature map $m^*$, which is shared across all edges, can then be written as:    
	\begin{equation}
	\begin{aligned}
	\bar{m}_{e}\left(x_{e}\right)=\frac{\exp \left(\alpha_{c o n v}\right)}{\exp \left(\alpha_{\operatorname{conv}}\right)+\exp \left(\alpha_{skip}\right)} o_{e}\left(x_{e}\right) +\\
	\frac{\exp \left(\alpha_{skip}\right)}{\exp \left(\alpha_{\operatorname{conv}}\right)+\exp \left(\alpha_{skip}\right)} x_{e}
	\end{aligned}
	\end{equation}
	where $\alpha_{conv}$ and $\alpha_{skip}$ are the architecture parameters defined in DARTS. $o_{e}(x_{e})$ is the output of the convolution operation, and $x_{e}$ is the skip connection (i.e., the input feature map of edge $e$). By minimizing $\operatorname{var}\left( \bar{m}_{e} - m^{*} \right)$, we can get the optimal $\alpha_{conv}^{*}$ and $\alpha_{skip}^{*}$ as:
	
	\begin{equation}
	\begin{aligned}
	&\alpha_{conv}^{*} \propto \operatorname{var}\left(x_{e}-m^{*}\right), \\
	&\alpha_{skip}^{*} \propto \operatorname{var}\left(o_{e}\left(x_{e}\right)-m^{*}\right).
	\end{aligned}
	\end{equation}
	During the training of the search network, $x_{e}$ will  be theoretically closer to $m^{*}$ than $o_{e}\left(x_{e}\right)$,  resulting in $\alpha_{skip}$ to be greater than $\alpha_{conv}$. This leads to failed search because cells are full of skip-connections. Therefore, the magnitude of architecture parameters, essentially, cannot indicate how much the operation contributes to the search network’s performance. 
	
	The above procedure shows that, in DARTS, the optimization of the evaluation network is separated from the search process of architectures, \emph{i.e.}, Eq.(\ref{Eq_darts_s}). As a consequence, the magnitude of architecture parameters cannot indicate how much the operation contributes to the search network’s performance, leading to the discovered architectures are sub-optimal for the final evaluation networks.
	
	\vspace{-0.cm}
	\subsection{Cyclic Differentiable Architecture Search}
	\vspace{-0.cm}
	Different from previous methods~\cite{DARTS,PDARTS,PCDARTS} where the search network is separated from the evaluation network, our CDARTS integrates the two networks into a unified architecture and models the architecture search as a joint optimization problem of the two networks:
	\vspace{-0.cm}
	% \begin{equation}
	% \centering
	% \begin{aligned}
	% \min_{\alpha}\quad & \mathcal{L}_{val} (w_E^*, w_S^*, \alpha),
	% \\
	% \mathrm{s.t.}\quad  w_{S}^* & = \mathop{\arg \min}_{w_S} \mathcal{L}_{train} (w_S, \alpha),
	% \\
	% w_{E}^* & = \mathop{\arg \min}_{w_E} \mathcal{L}_{val} (w_E, \alpha), 
	% \end{aligned}
	% \label{cdarts} 
	% \vspace{0.1cm}
	% \end{equation}
	\begin{equation}
	\centering
	\begin{aligned}
	\min_{\alpha}~~ & \mathcal{L}_{val} (w_E^*, w_S^*, \alpha),
	\\
	\mathrm{s.t.} & \left\{ 
	\begin{aligned}
	w_{S}^* & = \mathop{\arg \min}_{w_S} \mathcal{L}_{train} (w_S, \alpha), 
	\\
	w_{E}^* & = \mathop{\arg \min}_{w_E} \mathcal{L}_{val} (w_E, \alpha), 
	\end{aligned}
	\right.
	\end{aligned}
	\label{cdarts} 
	\end{equation}
	% \HL{(maybe use different notations for $\mathcal{L}_{val}$ in $\mathcal{L}_{val} (w_E^*, w_S^*, \alpha)$ to distinguish from $\mathcal{L}_{val} (w_E, \alpha)$)}
	where $w_E$ is the weight of the evaluation network. 
	To optimize this objective function, we propose an alternating learning algorithm, consisting of two iterative stages: a \emph{separate learning} stage and a \emph{joint learning} stage.
	The former is to train the search and evaluation networks on the \emph{train} and \emph{val} datasets respectively, and enable them to have good initialization weights $w_S$ and $w_E$ . The latter is to learn the architecture weights $\alpha$ and the network weights jointly. 
	These two stages are performed alternatively until convergence, leading to a joint optimization between the network search and evaluation, as presented in Algorithm~\ref{alg:search}. This cyclic process allows the evolution of searched architectures to fit the final target evaluation network. 
	
	\vspace{-0.cm}
	\textbf{Stage 1: Separate Learning.} The goal of this stage is to train the search and evaluation networks individually and make them adaptive to the given data. Specifically, for the search network, the architecture weights $\alpha$ are initialized randomly before training. Then, the weights $w_S$ are optimized on the $train$ data as follows:
	\vspace{-0.cm}
	\begin{equation}
	\centering
	\begin{aligned}
	w_S^* & = \mathop{\arg \min}_{w_S} \mathcal{L}^S_{train} (w_S, \alpha),
	\end{aligned}
	\label{eq:s} 
	\vspace{0.1cm}
	\end{equation}
	\noindent where $\mathcal{L}^S_{train}$ defines the loss function, and $w_S^*$ denotes the learned weight. For the image classification problem, we specify $\mathcal{L}^S_{train}$ as a cross-entropy loss.
	
	For the evaluation network, its internal cell structures are generated by discretizing the learned weights $\alpha$. Following the previous work~\cite{NASNet,DARTS}, we retain the top-$k$ ($k=2$) operations for each node in the cells by thresholding the learned values in $\alpha$. 
	The separate learning of the evaluation network is performed on the \emph{val} set through optimizing the following objective function:
	\begin{equation}
	\centering
	\vspace{-0.cm}
	\begin{aligned}
	{w_E^{\ast}} & = \mathop{\arg \min}_{w_E} \mathcal{L}^E_{val} (w_E, \overline{\alpha}),
	\end{aligned}
	\label{eq:e} 
	\vspace{0.1cm}
	\end{equation}
	where $\overline{\alpha}$ indicates the top-$k$ discretization of the continuous $\alpha$, and $w_S^*$ represents the learned weight. The separate learning of $w_S$ and $w_E$ enables the search and evaluation networks to obtain a good initialization.
	
	\floatname{algorithm}{Algorithm}
	\renewcommand{\algorithmicrequire}{\textbf{Input:}}
	\renewcommand{\algorithmicensure}{\textbf{Output:}}
	\setlength{\intextsep}{0.6cm}
	\setlength{\textfloatsep}{0.6cm}
	\begin{algorithm}[!tb]
		\vspace{-0.cm}
		\caption{Cyclic Differentiable Architecture Search}
		\begin{algorithmic}[1]
			\Require The \emph{train} and \emph{val} data, search and evaluation iterations $S_S$ and $S_E$, update iterations $S_U$, architecture weights $\alpha$, and weights $w_S$ and $w_E$ for search S-Net and evaluation 
			E-Net.
			
			\Ensure Evaluation network.
			\State Initialize $\alpha$ with randoms
			\State Initialize $w_S$
			
			\For{each search step $i \in [0, S_{S}]$}
			\vspace{1mm}
			\State /* Stage 1: Separate training */
			
			\If {$i \mathrm{~mod~} S_{U}$ = 0}
			\State Discretize $\alpha$ to $\overline{\alpha}$ by selecting the top-$k$ elements
			\State Generate E-Net with $\overline{\alpha}$ % and $w_S$
			\For{each evaluation step $j\in [0, S_{E}]$}
			\State Calculate $\mathcal{L}_{val}^E$ according to Eq.(\ref{eq:e})
			\State Update $w_{E}$
			\EndFor
			\EndIf
			
			\vspace{1mm}
			\State /* Stage 2: Joint training */
			\State Calculate $\mathcal{L}_{val}^{S}$, $\mathcal{L}_{val}^{E}$ and $\mathcal{L}_{val}^{S,E}$   according to Eq.(\ref{eq:joint})
			\State Jointly update $\alpha$ and $w_E$
			\State Calculate $\mathcal{L}_{train}^S$ according to Eq.(\ref{eq:s})
			\State Update $w_S$
			
			\EndFor
		\end{algorithmic}
		\label{alg:search}
	\end{algorithm}

\begin{figure}[t]
		\centering
		%\vspace{0.3cm}
		% \includegraphics[scale=0.95]{figures/arch.pdf}
		\includegraphics[width=0.5\textwidth]{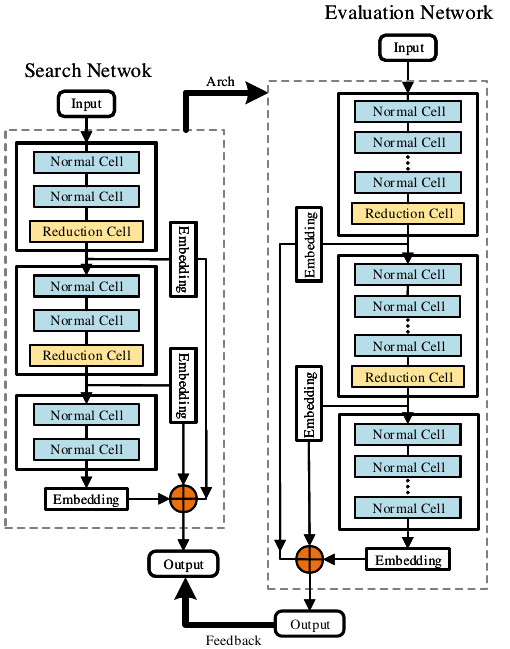}
		\caption{Illustration of the proposed CDARTS. CDARTS contains two networks, the search network (left) and the evaluation network (right). The \emph{Embedding} module maps each stage feature to a one-dimensional vector.}
		\label{fig:arch}
		\vspace{-1em} 
	\end{figure}

	\textbf{Stage 2: Joint Learning.} In this optimization stage, the search algorithm updates the architecture weights $\alpha$  with the feature feedback from the evaluation network through introspective distillation. More concretely, the joint optimization of the two networks is formulated as:
	\begin{equation}
	\centering
	\vspace{-0.cm}
	\begin{aligned}
	\alpha^*, w^*_E =  \mathop{\arg \min}_{\alpha, w_E} ~~  \mathcal{L}^S_{val}({w_S^*}, \alpha) +
	\mathcal{L}^E_{val}({w_{E}}, \overline{\alpha})&  \\
	+ \lambda~\mathcal{L}^{S, E}_{val}(w_S^*, \alpha,w_E,\overline{\alpha}) &,
	\end{aligned}
	\label{eq:joint}
	\vspace{0.1cm}
	\end{equation} 
	where minimizing $\mathcal{L}^S_{val}({w_S^*}, \alpha)$ is to optimize the architecture weights $\alpha$ with the fixed weights $w^{*}_S$ in the search network, $\mathcal{L}^E_{val} ({w_{E}}. \overline{\alpha})$ is to optimize the weights $w_E$ with a fixed architecture $\overline{\alpha}$ in the evaluation network. The $\mathcal{L}^{S, E}_{val}(w_S^*, \alpha,w_E,\overline{\alpha})$ represents the proposed \emph{introspective distillation}, which allows the knowledge transfer from the evaluation network to the search network. $\mathcal{L}^{S, E}_{val}(\cdot)$ employs the features derived from the evaluation network as a supervision signal to guide the updates of the architecture hyperparameter $\alpha$ in the search network. The introspective distillation is formulated by a soft target cross entropy function as:
	% the architecture hyperparameter
	\vspace{0.1cm}
	\begin{equation} \label{eq:kl}
	\centering
	\mathcal{L}^{S, E}_{val}(w_S^*,\alpha, w_E,\overline{\alpha}) = \frac{T^2}{N}\sum_{i=1}^{N}  (p(w_E,\overline{\alpha}) \log(\frac{ p(w_E,\overline{\alpha})}{q(w_S^*,{\alpha})} ) ),
	\vspace{0.1cm}
	\end{equation}
	where $N$ is the number of training samples, and $T$ is a temperature coefficient (set to $2$). Here, $p(\cdot)$ and $q(\cdot)$ represent the output feature logits of the evaluation and search networks respectively, each of which is calculated as the soft target distribution~\cite{hinton2015distilling} over the feature logits, \emph{i.e.}, 
	\vspace{0.1cm}
	\begin{equation}
	\centering
	\begin{aligned}
	p(w_E,\overline{\alpha}) &= \frac{\exp({f^E_i/T})}{\sum_j \exp({f^E_j/T}) }, \\
	q(w_S^*,{\alpha}) &= \frac{\exp({f^S_i/T})}{\sum_j \exp({f^S_j/T})},
	\end{aligned} 
	\vspace{0.1cm}
	\end{equation}
	where $f_i^E$ and $f_i^S$ denote the features generated by the search and evaluation networks (see Fig.~\ref{fig:arch}). The optimization of Eq. (\ref{eq:kl}) distills the knowledge of the evaluation network to guide the updates of architecture weights in the search network, while the joint training in Eq. (\ref{eq:joint}) allows the knowledge transfer between the two networks.
	
	Compared to the classical distillation method~\cite{hinton2015distilling}, CDARTS performs knowledge transfer in an introspective manner such that the search network can get the informative feedback from the evaluation network consecutively. The introspective distillation has three advantages. First, the teacher model is endogenous, which enables our method to adapt to any scenario. For example, there is no need to train an extra teacher, which greatly saves computing resources. Second, instead of learning from a fixed teacher model, our method updates the teacher models during the search process, thus the teacher model can adapt to the learning state of the student model. This has also been proved by Pham \etal~\cite{meta_pseudo} that when the teacher and student network are alternately optimized, the student network may learn from better distributions and achieve good validation performance. Last but not the least, the introspective distillation enables the search network to get the feedback information from the current search status, which prevents the search process from collapse~\cite{liang2019darts+} and improves the search stability.
	
	In addition, it has been observed that DARTS approaches tend to search the architectures with plenty of skip-connection operations rather than convolutions or poolings, because that the skip-connection allows rapid gradient descent~\cite{PDARTS,liang2019darts+}. This is essentially a kind of overfitting of architecture search. To address this issue, we impose an $\ell_1$-norm regularization on the weight of the skip-connection operation in the architecture weights $\alpha$ as:
	\vspace{0.1cm}
	\begin{equation}
	\centering
	\mathcal{L}_{reg} = \lambda'~\|\alpha\|_1,
	\label{eq:L1} 
	\vspace{0.1cm}
	\end{equation}
	where $\|\cdot\|_1$ represents the $\ell_1$ norm, and $\lambda'$ is a positive tradeoff coefficient. Eq. (\ref{eq:L1}) is finally optimized with Eq. (\ref{eq:joint}) jointly as an auxiliary item to inhibit overfitting.
	
	It is worth noting that during the separate learning of the evaluation network, \emph{i.e.}, Eq. (\ref{eq:e}), we use a weight-sharing strategy for updating weight $w_E$ to alleviate the insufficient training. Specifically, when the architecture weights $\overline{\alpha}$ are updated, the architecture of the evaluation network will be changed accordingly. The weights of the new evaluation network are initialized with the parameters inheriting from previous training. In other words, the evaluation network has a one-shot model that shares weights between architectures that have common edges. This speeds up the mature convergence of the new evaluation network, thus elevating its capacity on feature representation. This weight-sharing strategy is different from that in single-path one-shot approaches~\cite{randomNAS,spos}, which performs random sampling for architecture selection. In contrast, we select architectures to be optimized by the search network, which alleviates the issue of unbalanced training in prior methods ~\cite{spos,sciuto2019evaluating}.

	% \end{wrapfigure}
	%
	
	% \vspace{0.2cm}
	\subsection{Network Architecture}
	% \vspace{0.2cm}
	The network architecture of the proposed CDARTS is presented in Fig.~\ref{fig:arch}.
	It consists of two branches: a search network with a few stacked cells and an evaluation network with dozens of cells. Note that the search and evaluation networks share the same architectures with previous DARTS approaches~\cite{DARTS,PDARTS,PCDARTS,chu2019fairdarts}.
	
	% \begin{figure}[t]
	% 	\centering
	% 	%\vspace{0.3cm}
	% 	% \includegraphics[scale=0.95]{figures/arch.pdf}
	% 	\includegraphics[width=0.5\textwidth]{figures/arch.pdf}
	% 	\caption{Illustration of the proposed CDARTS. CDARTS contains two networks, the search network (left) and the evaluation network (right). The \emph{Embedding} module maps each stage feature to a one-dimensional vector.}
	% 	\label{fig:arch}
	% 	\vspace{-0.cm} 
	% \end{figure}

	For information transfer, we build up connections between the two branches. Concretely, there is an \emph{architecture transfer} path delivering the discovered cell motifs from the search branch to the evaluation branch, as the top bold arrow presented in Fig.~\ref{fig:arch}. Note that the cell structure discovered by the search network is a fully connected graph. In other words, all the candidate operations are applied to calculate the feature of each node in the graph, \emph{i.e.}, Eq~(\ref{Eq_darts_node}).
	When using this continuous cell structure to construct a new evaluation network, we need to conduct discretization first. Following previous works~\cite{DARTS}, we retain the top-$k$ ($k=2$) strongest operations among all the candidates from all the previous nodes. This derived discrete cell structure serves as the basic building block for the evaluation branch. 
	
	On the other hand, there is another \emph{introspective distillation} path transferring the feature feedback of the evaluation branch to the search branch, as the bottom solid arrow shown in Fig.~\ref{fig:arch}. The feedback serves as the supervision signal for the search network to find better cell structures. In details, we use multi-level features of the evaluation network as the feedback signals, since they are representative on capturing image semantics.  As the lateral \emph{embedding} connections presented in Fig.~\ref{fig:arch}, the multi-level features combine low-resolution, semantically strong features with high-resolution, semantically weak features. The features are derived from the outputs of each stage, and then passed through an \emph{embedding} module to generate the corresponding feature logits. The function of the \emph{embedding} module is to project the dense feature maps into a low dimensional subspace. The obtained logits of the evaluation network is used as the supervision signals for the search network through a soft cross-entropy layer, as in Eq.~(\ref{eq:kl}).% (the detailed function will be elaborated in the next subsection).

	\section{Experiments}
	\label{sec:exp}
	
	We evaluate the proposed CDARTS algorithm on the image classification task and conduct a series of experiments on CIFAR~\cite{krizhevsky2009learning}, ImageNet~\cite{krizhevsky2012imagenet}, as well as the recent proposed NATS-Bench benchmark~\cite{dong2021nats}.  Furthermore, we apply our introspective distillation search method into the chain-structured search space~\cite{Survey} to verify the generalization capability of our method.
	
	\vspace{-0.cm}
	\subsection{Implementation Details}
	\vspace{-0.cm}
	\label{sec:imdetail}
	
	\begin{table*}[!t]
		\small
		\begin{center}
			\centering
			\caption{
				%  used in both the search and the evaluation networks
				The hyperparameters used in search. BS, LR, OPT, MOME and WD are the batch size, learning rate, optimizer, momentum and weight decay. $\lambda$ and $\lambda'$ are the coefficients in Eq.~\ref{eq:joint} and Eq.~\ref{eq:L1}. Runs means the number of independent runs. The unit $s$ of $S_S$, $S_E$ and $S_U$ in Algorithm~\ref{alg:search} represents the number of steps in one epoch.
			}
			\label{table:settings}%	\scalebox{1.1}
			{
				\begin{tabular}{ c | c | c | c | c | c | c | c | c | c | c | c } 
					\toprule[1.2pt]
					Benchmark  & BS  & LR & OPT & MOME & WD & $S_S$  & $S_E$ & $S_U$ & $\lambda$ & $\lambda'$ &Runs   \\
					\hline
					NATS-Bench & 64 & 0.1 & SGD  & 0.9  & 3e-4   & 50$s$ & 1$s$ & 1$s$ & 4 & 5 & 3 \\
					CIFAR10  & 64 & 0.1  & SGD  & 0.9  & 3e-4   & 30$s$ & 1$s$ & 1$s$ & 4 & 5 & 5 \\
					ImageNet  & 1024 & 0.8  & SGD  & 0.9  & 3e-5  & 30$s$ & 3$s$ & 1$s$ & 4 & 5 & 5 \\
					\bottomrule[1.2pt]
				\end{tabular}
			}
		\end{center}
		\vspace{-1em}
	\end{table*}
	
	In Tab.~\ref{table:settings}, we elaborate the setting of the hyperparameters used in CDARTS on different benchmarks. The other settings are the same as DARTS families~\cite{DARTS,PDARTS,PCDARTS}. In line with prior works~\cite{DARTS,PDARTS,PCDARTS}, the architectures found on CIFAR10 and ImageNet need to be trained from scratch.
	
	\textbf{CIFAR10.} To train the networks, we divide the $50$K training images of CIFAR10~\cite{krizhevsky2009learning} into two equal parts. One part serves as the \emph{train} set to learn the weight $w_S$ of the search network, while the other part works as the \emph{val} set to optimize the architecture hyperparameter $\alpha$ and the evaluation network weights $w_E$. %\BLUE{In the search phase, $w_{E}^{*}$ is only optimized on the validation set. When the search phase is over, we retrain the discovered architectures on both train+validation set for a fair comparison with other DARTS methods.}
	In the search phase, $w_{E}^{*}$ is only optimized on the validation set. After the search, we retrain the discovered architectures on both train and validation sets for a fair comparison with other DARTS methods.
	During training, the total number of search step $S_S$ is set to $30$ epochs,
	the evaluation step $S_E$ and the update step $S_U$ are both set to $1$ epoch.
	When training the weight $w_S$ and $w_E$ individually, the batch size, learning rate, momentum and weight decay are set to $64$, $0.1$, $0.9$ and $3\times10^{-4}$, respectively. When jointly updating $\alpha$ and $w_E$, we adopt the Adam optimizer~\cite{kingma2014adam} with a fixed learning rate of $3\times10^{-4}$. The momentum and weight decay are set to $\{0.5,0.99\}$ and $3\times10^{-4}$, respectively. The coefficient $\lambda$ is fixed to $4$, while $\lambda'$ decays from $5$ to $0$ in the first $20$ epochs and remains at $0$ in the last $10$ epochs. 
	
	\textbf{ImageNet.} ImageNet~\cite{krizhevsky2012imagenet} is much larger than CIFAR~\cite{krizhevsky2009learning} in both scale and complexity.
	In line with prior works, \emph{i.e.}, PCDARTS~\cite{PCDARTS} and DARTS+~\cite{liang2019darts+}, 
	we randomly sample 10\% images from ImageNet to form the \emph{train} data, while sampling another 10\% images to form the \emph{val} data. 
	The construction of networks tested on ImageNet is similar to the one on CIFAR, but has two differences. First, to align with previous works~\cite{PDARTS,DARTS,PCDARTS} on ImageNet, we set the number of cells to $8$ and $14$ for the search and evaluation networks. Second, following the same setting as DARTS~\cite{DARTS,PDARTS,PCDARTS}, the stem layer contains $3$ convolution operations reducing the feature size from 224$\times$244 to 28$\times$28. 
	During search, we first pre-train the search network with 10 epochs when the architecture hyperparameters are fixed. 
	Then, the number of search step $S_S$ is set to $30$ epochs, while the evaluation step $S_E$ is set to $3$ epochs. When training the weight $w_S$ and $w_E$ individually, the batch size, learning rate, momentum and weight decay are set to $1024$, $0.8$, $0.9$ and $3\times10^{-5}$, respectively. When jointly updating $\alpha$ and $w_E$, we adopt the Adam optimizer~\cite{kingma2014adam} with a fixed learning rate of $3\times10^{-4}$. The momentum and weight decay are set to $\{0.5,0.99\}$ and $3\times10^{-5}$, respectively. The coefficient $\lambda$ is fixed to $4$, and $\lambda'$ decays from 5 to 0 in the first $20$ epochs and remains at $0$ in the rest epochs. We use 8 NVIDIA Tesla V100 GPUs with a batch size of $1024$ to search on ImageNet. It takes about 5 hours with our PyTorch~\cite{paszke2019pytorch} implementation.
	
	\vspace{-0.cm}
	\subsection{Evaluation on NAS Benchmark}
	\vspace{-0.cm}
	
	\begin{table*}[t]
		\centering
		\small

		\caption{Comparison with 10 NAS methods provided by NATS-Bench benchmark~\cite{dong2021nats} topology search space $\textbf{\textit{S}}_{t}$. %The mean $\pm$ std of 3 independent runs with different random seeds are reported.
			\textbf{Optimal} indicates the best performing architecture in the search space.
		}
		% The \emph{mean $\pm$ std} of 3 independent runs are reported.
		%\resizebox{\textwidth}{!}{%
		\begin{tabular}{ c   c   c   c   c   c   c   c }
			% \arrayrulecolor{blue}
			\toprule[1.2pt]
			\multirow{2}{*}{Method} & \multicolumn{2}{c}{CIFAR10} & \multicolumn{2}{c}{CIFAR100} & \multicolumn{2}{c}{ImageNet-16-120} \\
			\cline{2-7}
			& validation     & test           &   validation   &   test    & validation & test  \\
			\bottomrule[0.7pt]
			ResNet~\cite{he2016deep}           & 90.83 & 93.91 & 70.50 & 70.89 & 44.10 & 44.23  \\ %\cline{1-1}\cline{3-8}
			\textbf{Optimal} & 91.61 & 94.37(94.37) & 73.49 & 73.51(73.51) & 46.73 & 46.20(47.31)  \\
			\hline
			%REA       &  0.02 & 91.04$\pm$0.47 & 93.81$\pm$0.45 & 71.41$\pm$1.33 & 71.50$\pm$1.24 & 44.79$\pm$1.27 & 45.10$\pm$1.47 \\
			REA~\cite{real2019regularized}      & \RED{91.25$\pm$0.31} & \RED{94.02$\pm$0.31} & \RED{72.28$\pm$0.95} & \RED{72.23$\pm$0.84} & \RED{45.71$\pm$0.77} & \RED{45.77$\pm$0.80} \\
			REINFORCE~\cite{REINFORCE} & 91.12$\pm$0.25 & 93.90$\pm$0.26 & 71.80$\pm$0.94 & 71.86$\pm$0.89 & 45.37$\pm$0.74 & 45.64$\pm$0.78 \\
			RANDOM~\cite{RS}       & 91.07$\pm$0.26 & 93.86$\pm$0.23 & 71.46$\pm$0.97 & 71.55$\pm$0.97 & 45.03$\pm$0.91 & 45.28$\pm$0.97 \\
			%REINFORCE &  0.12 & 89.81$\pm$0.92 & 92.99$\pm$0.96 & 69.75$\pm$1.62 & 69.85$\pm$1.64 & 43.15$\pm$2.42 & 43.40$\pm$2.53 \\
			BOHB~\cite{BOHB}      & 91.17$\pm$0.27 & 93.94$\pm$0.28 & 72.04$\pm$0.93 & 72.00$\pm$0.86 & 45.55$\pm$0.79 & 45.70$\pm$0.86 \\
			%BOHB      &  3.59 & 90.72$\pm$0.58 & 93.50$\pm$0.55 & 70.56$\pm$1.32 & 70.65$\pm$1.31 & 44.13$\pm$1.43 & 44.18$\pm$1.60 \\
			\hline
			ENAS~\cite{ENAS}   & 90.20$\pm$0.00 & 93.76$\pm$0.00 & 70.21$\pm$0.71 & 70.67$\pm$0.62 & 40.78$\pm$0.00 & 41.44$\pm$0.00 \\
			 
		    SPS~\cite{randomNAS}   & 87.60$\pm$0.61 & 91.05$\pm$0.66 & 68.27$\pm$0.72 & 68.26$\pm$0.96 & 39.37$\pm$0.34 & 40.69$\pm$0.36 \\
		     
			SETN~\cite{SETN}    & 90.02$\pm$0.97 & 92.72$\pm$0.73 & 69.19$\pm$1.42 & 69.36$\pm$1.72 & 39.77$\pm$0.33 & 39.51$\pm$0.33 \\
			\cdashline{1-7}[0.8pt/3pt]
			DARTSV1~\cite{DARTS} & 49.27$\pm$13.44 & 59.84$\pm$7.84 & 61.08$\pm$4.37 & 61.26$\pm$4.43 & 38.07$\pm$2.90 & 37.88$\pm$2.91 \\
			DARTSV2~\cite{DARTS} & 58.78$\pm$13.44 & 65.38$\pm$7.84 & 59.48$\pm$5.13 & 60.49$\pm$4.95 & 37.56$\pm$7.10 & 36.79$\pm$7.59 \\
			GDAS~\cite{GDAS}   & 89.69$\pm$0.72 & 93.23$\pm$0.58 & 68.35$\pm$2.71 & 68.17$\pm$2.50 & 39.55$\pm$0.00 & 39.40$\pm$0.00 \\
			\textbf{CDARTS}  & \BLUE{91.12$\pm$0.44} & \RED{94.02$\pm$0.31} & \BLUE{72.12$\pm$1.23} & \BLUE{71.92$\pm$1.30} & 45.09$\pm$0.61 & 45.51$\pm$0.72 \\
			%\textbf{optimal}&  N/A &   N/A      & N/A         & 6111    & 9930  & 9930    & 10676 & 857  \\
			\bottomrule[1.2pt]
		\end{tabular}
		%}
		\label{table:benchmarking}
		\vspace{-0.cm}
	\end{table*}
	
	\begin{figure*}[t]
		\vspace{-0.25em}
		\centering
		\caption{Test and validation accuracy ($mean\pm std$) \emph{v.s.} search epochs on NATS-Bench benchmark~\cite{dong2021nats} topology search space $\textbf{\textit{S}}_{t}$. DARTSV1 and DARTSV2 indicate the first-order and second-order DARTS methods~\cite{DARTS}, whose results are provided by the benchmark~\cite{dong2021nats}.}
		\includegraphics[width=\linewidth]{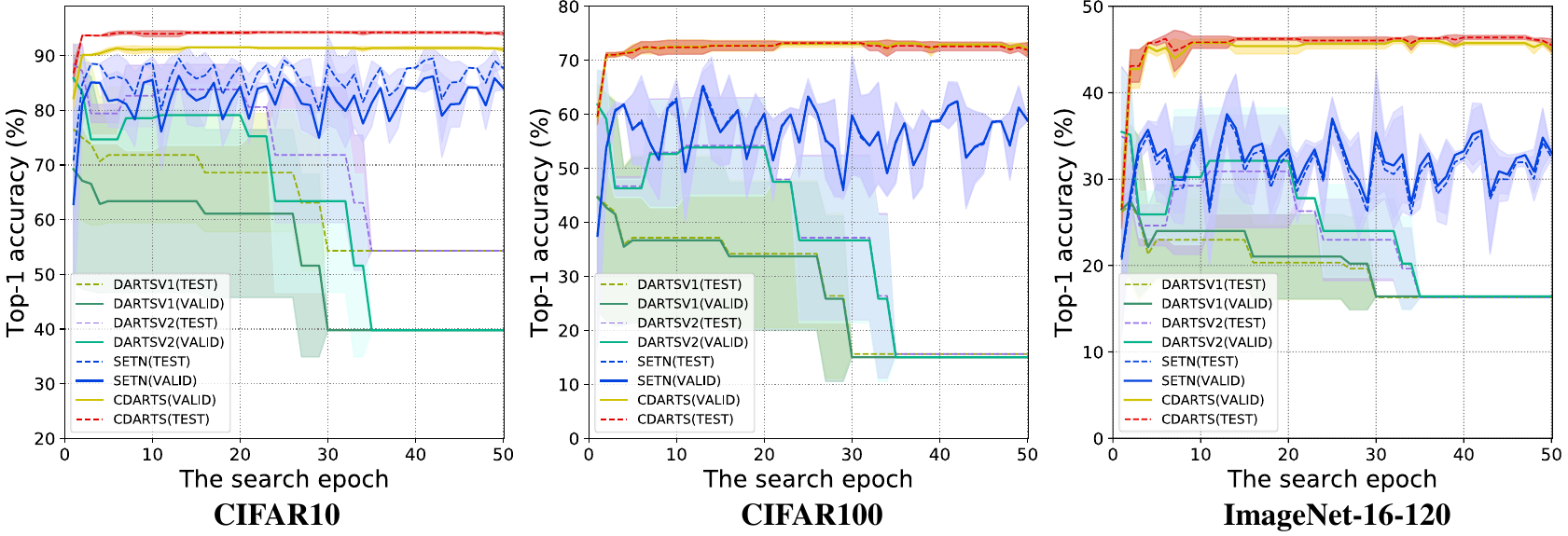}
		% 	\vspace{-1em}
		
		\label{fig:bench201}
		\vspace{-0.5em}
	\end{figure*}
	
		\begin{table}[t]
		\renewcommand{\arraystretch}{1.3}
		\centering
		\small
		\caption{Top-1 accuracy of searched architectures in five independent search runs on CIFAR10.}
		\label{table:c10runs}
		\scalebox{0.87}
		{	
			\begin{threeparttable}[b]
				% \resizebox{0.6\textwidth}{!}{
				\begin{tabular}{lccccccc}	
					\toprule[1.2pt]
					\multirow{2}*{\textbf{Methods}} & \multicolumn{5}{c}{\textbf{Runs}} & \multirow{2}*{\textbf{ Mean$\pm$std}} \\
					\cline{2-6} & \textbf{\#1} & \textbf{\#2} & \textbf{\#3} & \textbf{\#4} & \textbf{\#5} \\
					\toprule[1.2pt]
					DARTSV1(\%) &97.11 &96.85 &97.01 &96.93 &96.73 &96.93$\pm$0.13 \\
					DARTSV2(\%) &96.89 &96.32 &97.23 &96.86 &96.94 &96.85$\pm$0.29 \\
					PCDARTS(\%) &97.28 &97.33 &97.43 &97.25 &97.36 & 97.33$\pm$0.07 \\
					\textbf{CDARTS}(\%) &\textbf{97.53} &\textbf{97.45} &\textbf{97.60} &\textbf{97.52} &\textbf{97.54} & \textbf{97.52$\pm$0.04} \\
					\bottomrule[1.2pt]		
				\end{tabular}
			\end{threeparttable}
		}
		\vspace{-0.5em}
	\end{table}
	
	NATS-Bench benchmark~\cite{dong2021nats} includes the search space of 15,625 neural cell candidates for architecture topology and 32,768 for architecture size on three datasets. We conduct our experiments on the topology search space $\textbf{\textit{S}}_{t}$,  covering all possible architectures generated by the fixed search space of 4 nodes and 5 associated operation options. It evaluates all the architectures on CIFAR10~\cite{krizhevsky2009learning}, CIFAR100~\cite{krizhevsky2009learning} and ImageNet-16-120~\cite{chrabaszcz2017downsampled}, and provides the corresponding performance. % of each architecture candidate. 
	It also benchmarks 13 recent NAS algorithms on the search space using the same setup for a fair comparison.
	
	According to the evaluation rules of NATS-Bench benchmark~\cite{dong2021nats}, \emph{i.e.}, \emph{no regularization for a specific operation} and \emph{report results of multiple searching runs}, we remove the $\ell_1$-norm regularization imposed on skip-connections, \emph{i.e.}, Eq.(\ref{eq:L1}), and report the results of three independent runs. All the DARTS-based methods perform $50$ searching epochs, following the same setting as~\cite{dong2021nats}.
	
	We first compare our method with the 10 NAS methods that have been benchmarked on NATS-Bench benchmark~\cite{dong2021nats}, such as the evolution algorithm based REA~\cite{real2019regularized} and the one-shot based ENAS~\cite{ENAS}. As presented in Tab.~\ref{table:benchmarking}, our CDARTS outperforms 9/10 methods on the three datasets. It achieves comparable performance to the REA method~\cite{real2019regularized}, while searching much faster. Both of REA and CDARTS perform superior to the human-design ResNet~\cite{he2016deep}, being close to theoretical optimum on the benchmark, \emph{i.e.}, the ``Optimal'' in Tab.~\ref{table:benchmarking}.  The original DARTS underperforms due to overfitting~\cite{dong2021nats}, and its standard deviation of performance is $0$ because the final found architecture has plenty of skip-connections. The underlying reason is that DARTS tends to overfit to the architectures with many skip-connection operations. GDAS~\cite{GDAS} is another DARTS based method. It performs much better than DARTS on NATS-Bench benchmark because it introduces an architecture sampling optimization algorithm to prevent overfitting. In contrast, our CDARTS can perform well-balanced and achieves superior performance. The underlying reason is due to the proposed unified architecture, which allows the search network to discover architectures tailored for the target evaluation network. Except for CDARTS, the performance of all the NAS approaches in Tab.~\ref{table:benchmarking} are provided by the official NATS-Bench benchmark~\cite{dong2021nats}. 
	
	% We simply reuse the provided results for a fair comparison. 

	We conduct another experiment to evaluate the searching stability.
	With the NATS-Bench benchmark~\cite{dong2021nats} benchmark, it is easy to track the validation and test accuracy of every discovered architecture after each training epoch, 
	as visualized in Fig.~\ref{fig:bench201}. We can see that DARTS~\cite{DARTS} achieves a relatively high accuracy at early stage, however, as the search process continues, its accuracy drops significantly and the stability becomes weak. Finally, DARTS falls into the overfitting status, in which the final architecture contains many skip-connections. In contrast, the performance of our CDARTS is improved gradually along with the increase of search epochs, and eventually reaches stable and converged. 
	This may attribute to the knowledge transfer between the search and evaluation networks. CDARTS continually leverages the supervision from the evaluation network to guide the search process, thus preventing the searched architecture from collapsing. Moreover, it is observed that the one-shot NAS method SETN~\cite{SETN} performs inferiorly to our method  in term of both accuracy and searching stability.
	
		\begin{figure*}[!t]
		\centering
		\vspace{0.cm}
		\caption{Cells discovered on CIFAR10 and ImageNet. ImageNet cells are deeper than CIFAR10.}
		\includegraphics[width=0.9\textwidth]{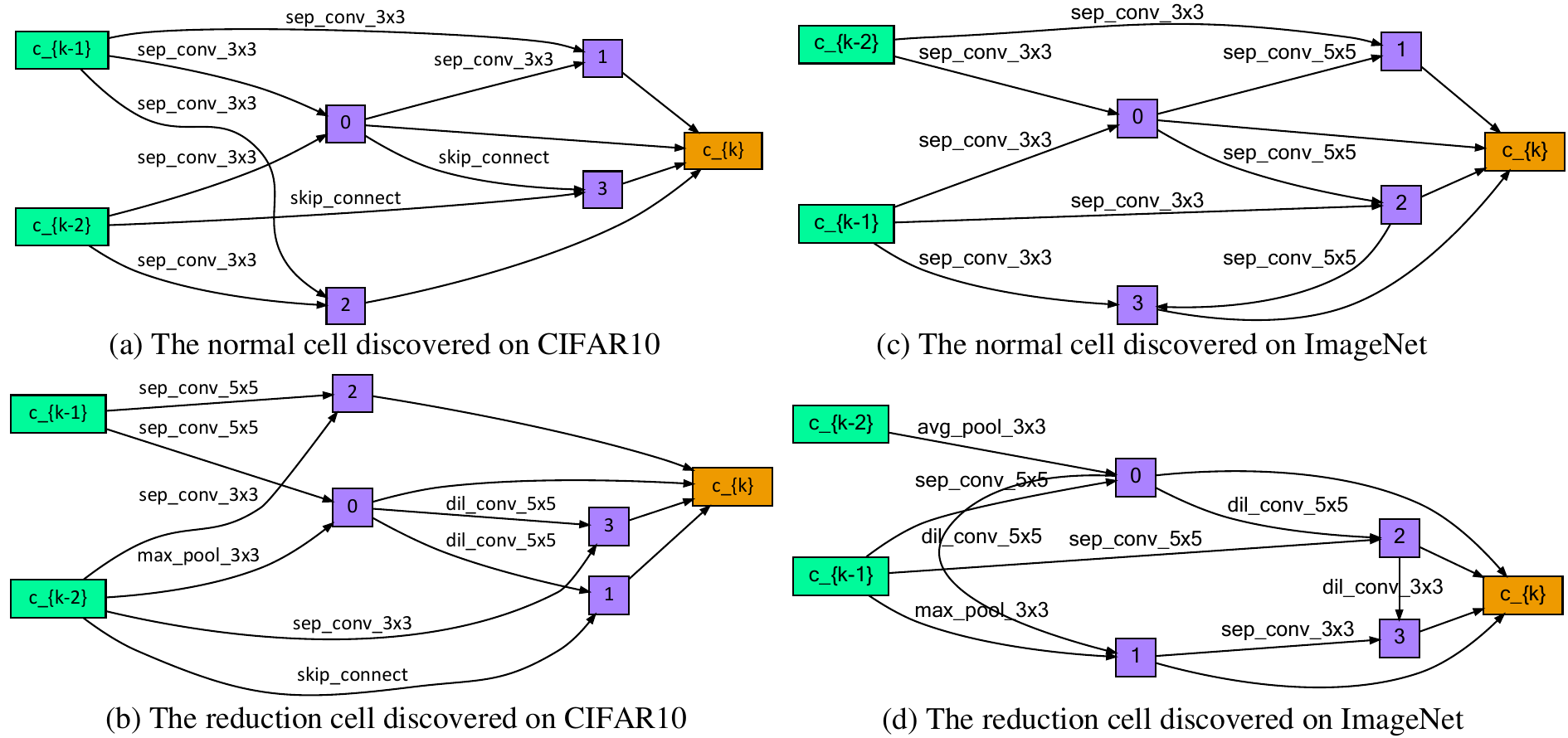}
		% \caption{Discovered normal (left) and reduction (right) cells on ImageNet.}
		\label{fig:imgcell} 
		\vspace{-0.25cm}
	\end{figure*}
	
	\begin{table*}[ht]
		\vspace{-0.cm}
		\small
		\centering
		\caption{Comparison with SOTA architectures on CIFAR10 and CIFAR100. $^\dagger$ indicates using the DARTS~\cite{DARTS} search space. Latency is  measured on an NVIDIA Tesla-P100 GPU with a batch size of 32 and an input size of 32$\times$32. }
		\vspace{-0.cm}
		\scalebox{1.}{
			\begin{tabular}{lcccccc}
				% \arrayrulecolor{blue}
				\toprule[1.2pt]
				\multicolumn{1}{l}{\multirow{2}{*}{Architecture}}  & \multicolumn{2}{c}{Test Acc. (\%)} & Param & Latency & Search Cost & Search \\
				\cline{2-3}
				\multicolumn{1}{c}{}  & CIFAR10 & CIFAR100 & (M) & (ms) & (GPU days) & \multicolumn{1}{c}{Method} \\
				\toprule[1.2pt]
				Wide ResNet~\cite{zagoruyko2016wide}  & $96.20$ & $82.70$ & $36.5$ & - & - & manual \\
				ResNeXt-29, 16x64d~\cite{xie2017aggregated}  & $96.42$ & $82.69$ & $68.1$ & - & - & manual \\
				DenseNet-BC~\cite{huang2017densely} & $96.52$ & $82.82$ & $25.6$ & - & - & manual \\
				\hline
				NASNet-A~\cite{NASNet}  & $97.35$ & -  & $3.3$ & - & $1800$ & RL \\
				% AmoebaNet-A + cutout & $3.34 \pm 0.06$ & - & $3.2$ & $3150$ & evolution \\
				AmoebaNet-B~\cite{real2019regularized}  & $97.45$ & - & $2.8$ & - & $3150$ & evolution \\
				PNAS~\cite{PNAS}  & $96.59$ & - & $3.2$ & - & $225$ & SMBO \\
				ENAS~\cite{ENAS}  & $97.11$ & - & $4.6$ & - & $0.5$ & RL \\
				NAONet~\cite{luo2018neural} & $96.82^{1}$ & \RED{$84.33$} & $10.6$ & - & $200$ & NAO \\
				\hline
				SNAS (moderate)~\cite{xie2018snas} & $97.15\pm0.02$ & - & $2.8$ & $30.2$ & $1.5$ & gradient \\
				ProxylessNAS~\cite{cai2018proxylessnas} & \RED{$97.92$} & - & $5.7$ & - & $4$ & gradient \\
				GDAS~\cite{GDAS} & $97.07$ & $81.62$ & $3.4$ & $30.6$ & $4$ & gradient \\
				\cdashline{1-6}[0.8pt/2pt]
				Random$^\dagger$ & $96.75\pm0.18$ &  - & $3.4$ & - & - & - \\
				DARTSV1~\cite{DARTS}$^\dagger$ & $97.00\pm0.14$ & $82.24$ & - & $3.3$ & $1.5$ & gradient \\
				DARTSV2~\cite{DARTS}$^\dagger$  & $97.24 \pm 0.09$ & $82.46$ & $40.9$ & $3.3$ & $4.0$ & gradient \\
				PDARTS~\cite{PDARTS}$^\dagger$  & $97.50$ & $83.45$ & $3.4$ & $40.9$ & $0.3$ & gradient \\
				PCDARTS~\cite{PCDARTS}$^\dagger$  & $97.43 \pm 0.06$ & - & $3.6$ & $40.7$ & $0.1$ & gradient \\
				FairDARTS~\cite{chu2019fairdarts}$^\dagger$  & $97.41 \pm 0.14$ & - & $3.8$ & - & $0.1$ & gradient \\
				LA-DARTS~\cite{xu2020latency}$^\dagger$  & $97.28 \pm 0.05$ & - & $2.7$ & $28.4$ & $0.7$ & gradient \\
				\textbf{CDARTS}$^\dagger$  & \BLUE{$97.52 \pm 0.04$} & \BLUE{$84.31$} & $3.9\pm0.08$ & $41.2\pm0.5$ & $0.3$ & gradient \\
				% \textbf{CDARTS-BIG(Ours)}$^\dagger$ & CIFAR10 & $1.68$ & $12.99$ & $7.4$ & $0.3$ & gradient \\
				\bottomrule[1.2pt]
			\end{tabular}
		}
		\label{tab:cifar_results}
		\vspace{-0.cm}
	\end{table*}

	\vspace{-0.cm}
	\subsection{Evaluation on CIFAR}
	\vspace{-0.cm}

	The CIFAR image classification dataset contains two subsets, CIFAR10 and CIFAR100. CIFAR10 has 10 classes. Each class consists of 6,000 images,
	in which 5,000 images are used for training and 1,000 for testing. CIFAR100 consists of 100 classes. Each class containing 600 images, where 500 images are used for training and the rest for testing.

	We search the architecture on CIFAR10 test set, and evaluate it on both CIFAR10 and CIFAR100. The cell topologies discovered on CIFAR10 are shown in Fig.~\ref{fig:imgcell}.  After discovering the evaluation network architecture, we retrain it by following the same setting as PDARTS~\cite{PDARTS}.
	Specifically, the number of channels is set to 36 and the structure is the same with the search stage. We use the entire CIFAR training images to train the network from scratch with $600$ epochs. To speed up the training, the batch size is set to $128$ and the learning rate decays from $0.025$ to $0$ with a cosine annealing~\cite{loshchilov2016sgdr}. We choose SGD~\cite{SGD} as the optimizer with a momentum of 0.9 and weight decay of $5 \times 10^{-4}$. In line with PDARTS, the drop-path~\cite{larsson2016fractalnet} rate is set to 0.3, the auxiliary towers~\cite{simonyan2014very} is set to 0.4, and the cutout regularization~\cite{devries2017improved} length is set to 16. The experiments are executed on one NVIDIA Tesla V100 GPU.

	\begin{table*}[ht]
		% \vspace{-0.3cm}
		\centering
		\small
		\caption{Results on ImageNet. $^\dag$ We use the same search space as DARTS~\cite{DARTS}. \textbf{MS} denotes mobile setting. $^*$ denotes use 10\% of ImageNet data for searching.} %\BLUE{$^\ddag$ 5 independent search runs on ImageNet and report the mean and standard deviation.}}
		% \resizebox{1.0\textwidth}{!}{%
		\vspace{-0.cm}
		\begin{tabular}{lcccccc}
			\toprule[1.2pt]
			%\hline
			\multirow{2}{*}{Architecture} & \multicolumn{2}{c}{Test Acc. (\%)} & Params  & $\times +$  & Search Cost & \multirow{2}{*}{Search Method} \\
			\cline{2-3}
			& Top-1 & Top-5 & (M) & (M) & (GPU days)  &  \\
			%\hline
			\toprule[1.2pt]
			Inception-V1~\cite{szegedy2015going} & $69.8$ & $89.9$ & $6.6$ & $1448$ & - & manual \\
			SqueezeNext~\cite{gholami2018squeezenext} & $67.5$ & $88.2$ & $3.23$ & $708$ & - & manual \\
			MobileNet-V2 ($1.4\times$)~\cite{sandler2018mobilenetv2} & $74.7$ & - & $6.9$ & $585$ & - & manual \\
			ShuffleNet-V2 ($2\times$)~\cite{shufflenetv2} & $74.9$ & - & $7.4$ & $591$ & - & manual \\
			\hline
			NASNet-A~\cite{NASNet} & $74.0$ & $91.6$ & $5.3$ & $564$ & $1800$ & RL \\
			AmoebaNet-C~\cite{real2019regularized} & $75.7$ & $92.4$ & $6.4$ & $570$ & $3150$ & evolution \\
			PNAS~\cite{PNAS} & $74.2$ & $91.9$ & $5.1$ & $588$ & $225$ & SMBO \\
			EfficientNet-B0~\cite{tan2019efficientnet} & \RED{$77.1$} & \RED{$93.2$} & $5.3$ & $390$ & - & RL \\
			MnasNet-92~\cite{tan2019mnasnet} & $74.8$ & $92.0$ & $4.4$ & $388$ & - & RL \\
			 
			SPOS~\cite{spos} & $74.3$ & - & - & $319$ & - & evolution \\
			 
			FairNAS-A~\cite{fairnas} & $75.3$ & $92.4$ & $4.6$ & $388$ & - & evolution \\
			 
			MobileNet-V3~\cite{mobilenetv3} & \BLUE{$76.6$} & - & $7.5$ & $356$ & - & RL \\
			 
			MoGA-A~\cite{moga} & $75.9$ & $92.8$ & $5.1$ & $304$ & $12$ & evolution \\
			\hline
			SNAS (mild)~\cite{xie2018snas} & $72.7$ & $90.8$ & $4.3$ & $522$ & $1.5$ & gradient \\
			XNAS ~\cite{nayman2019xnas} & $76.0$ & $ - $ & $5.2$ & $ - $ & $ 0.3 $ & gradient \\
			ProxylessNAS~\cite{cai2018proxylessnas} & $75.1$ & $92.5$ & $7.1$ & $465$ & $8.3$ & gradient \\
			ASAP~\cite{noy2019asap} & $73.3$ & - & - & - & $0.2$ & gradient \\
			\cdashline{1-7}[0.8pt/2pt]
			DARTS~\cite{DARTS}$^\dag$ & $73.3$ & $91.3$ & $4.7$ & $574$ & $4.0$ & gradient \\
			FairDARTS~\cite{chu2019fairdarts}$^\dag$& $75.6$ & $92.6$ & $4.3$ & $440$ & $3.0$ & gradient \\
			PDARTS~\cite{PDARTS}$^*\dag$ & $75.6$ & $92.6$ & $4.9$ & $557$ & $0.3$ & gradient \\
			PCDARTS~\cite{PCDARTS}$^*\dag$ & $75.8$ & $92.7$ & $5.3$ & $597$ & $3.8$ & gradient \\
			EnTranNAS (ImageNet)~\cite{yang2021towards}$^*\dag$ & $75.7$ & $92.8$ & $5.5$ & $637$ & $1.9$ & gradient \\
			% \textbf{CDARTS(CIFAR10)$^\dag$} & $75.6$ & $92.5$ & $5.4$ & $639$ & $0.3$ & gradient \\
			%\textbf{CDARTS$^\dag$} & \BLUE{$76.3\pm0.3$} & \BLUE{$92.9\pm0.2$} & $6.1$ & $732$ & $1.7$ & gradient \\
			\textbf{CDARTS(MS)$^*\dag$} & $75.9$ & $92.6$ & $5.4$ & $571$ & $1.7$ & gradient \\
			\textbf{$\text{CDARTS}^*\dag$} & $76.3\pm0.3$ & \BLUE{$92.9\pm0.2$} & $6.1\pm0.2$ & $701\pm32$ & $1.7$ & gradient \\
			%\textbf{CDARTS(BIG)$^\dag$} & $18.9$ & $5.0$ & $32.3$ & $5584$ & $1.7$ & gradient \\
			\bottomrule[1.2pt]
		\end{tabular}
		%}
		\label{tab:imagenet_results}
		\vspace{-0.2cm}
	\end{table*}

	We report the performance of five individual runs of different search algorithms in Tab.~\ref{table:c10runs}. 
	%CDARTS are about 0.2\% superior than PCDARTS and much better than DARTS. 
	In the five independent runs, CDARTS consistently outperforms prior DARTS~\cite{DARTS} and PCDARTS~\cite{PCDARTS}.
	Moreover, in terms of performance deviations, CDARTS performs better than prior methods, \emph{i.e.}, CDARTS: $97.52 \pm 0.04\%$ \emph{v.s.} DARTSV1~\cite{DARTS}: $96.93 \pm 0.13\%$, DARTSV2: $96.85 \pm 0.29\%$, PC-DARTS: $ 97.33 \pm 0.07\%$.
	%the architectures found by DARTS~\cite{DARTS} and PCDARTS~\cite{PCDARTS} in different runs suffer much higher standard deviations than that of our approach(DARTSV1: $\pm 0.13\%$, DARTSV2: $\pm 0.29\%$, PC-DARTS: $\pm 0.07\%$, CDARTS: $\pm 0.04\%$). 
	The cell motifs discovered on CIFAR-10 are presented in Fig.~\ref{fig:imgcell}.  We observe that the network prefers to choose separable convolutions~\cite{howard2017mobilenets}, and is therefore capable of improving the model capacity and serves as a key component for network construction. Besides, these cells usually contain either one or two skip-connections, and the depth of these cells are usually two (the longest path from input to output node).
	
	The performance of discovered cells are reported in Tab.~\ref{tab:cifar_results}. 
	%\BLUE{The result of random sampled cells on CIFAR10 is $96.75\pm 0.18$, which have a relative high standard deviation.}
	It is observed that our CDARTS achieves superior performance to some ConvNets designed manually or automatically. For instance, CDARTS surpasses the manually designed DenseNet-BC~\cite{huang2017densely} by $1.18$ and $3.93$ points on CIFAR10 and CIFAR100, respectively. Compared with the original DARTS method~\cite{DARTS}, CDARTS also achieves better performance ($97.52\pm 0.04$\% \emph{v.s.} $97.00\pm 0.14$\%).
	%while searching 5 times faster ($0.3$ \emph{v.s} $1.5$ GPU days)
	CDARTS is also slightly superior to the recently proposed PDARTS~\cite{PDARTS} and PCDARTS~\cite{PCDARTS} on both CIFAR10 and CIFAR100. It is worthy noting that the performance of ProxylessNAS~\cite{cai2018proxylessnas} is slightly better than CDARTS, but at the cost of a larger model parameter size and longer model search time. Moreover, compared with other non-gradient based methods, such as reinforcement learning (RL) or evolutionary algorithms, our method is also competitive. % our CDARTS still obtains relatively competitive performance. 
	For example, the RL-based ENAS~\cite{ENAS} method achieves $97.11$\% test accuracy on CIFAR10, which is slightly inferior to $97.52\%$ of CDARTS.

	%\begin{figure*}[!th]
	%	\centering
	%	\vspace{-0.cm}
	%	%\includegraphics[scale=0.4]{figures/5c10.pdf}
	%	\includegraphics[width=0.9\textwidth]{figures/c10_cells.pdf}
	%	% \caption{Cells discovered on CIFAR10 and ImageNet. ImageNet cells are more deeper than CIFAR10.}
	%	\caption{Discovered normal (left) and reduction (right) cells on CIFAR10.}
	%	\label{fig:c10cell} 
	%	\vspace{-0.cm}
	%\end{figure*}

	\vspace{-0.0cm}
	\subsection{Evaluation on ImageNet}
	\vspace{-0.0cm}
	ImageNet~\cite{krizhevsky2012imagenet} is a large-scale image classification dataset. 
	It consists of 1,000 categories with 1.2 million training images and 50K validation images. 
	Note that, for a fair comparison, the images in ImageNet are resized to 224$\times$224 pixels in our experiments in line with prior DARTS works~\cite{DARTS,PDARTS,PCDARTS,chu2019fairdarts}. 
	Once the search process is completed, we retrain the discovered architecture following the same setting as PDARTS. 
	The final evaluation network contains 14 cells with a channel number of 48. We retrain it for 250 epochs from scratch with full ImageNet training data. A standard SGD~\cite{SGD} optimizer is adopted with a momentum of 0.9 and weight decay of $3 \times 10^{-5}$. The learning rate is set to 0.5 with the cosine scheduler; meanwhile, a learning rate warmup~\cite{goyal2017accurate} is applied in the first 5 epochs. Same as DARTS, the label smoothing~\cite{szegedy2016rethinking} ratio is set to 0.1 and the auxiliary tower is set to 0.4.

	\begin{figure*}[!t]
		\centering
		% 	\vspace{0.1cm}
		\begin{minipage}{.45\textwidth}
			\centering
			\includegraphics[height=0.7\textwidth]{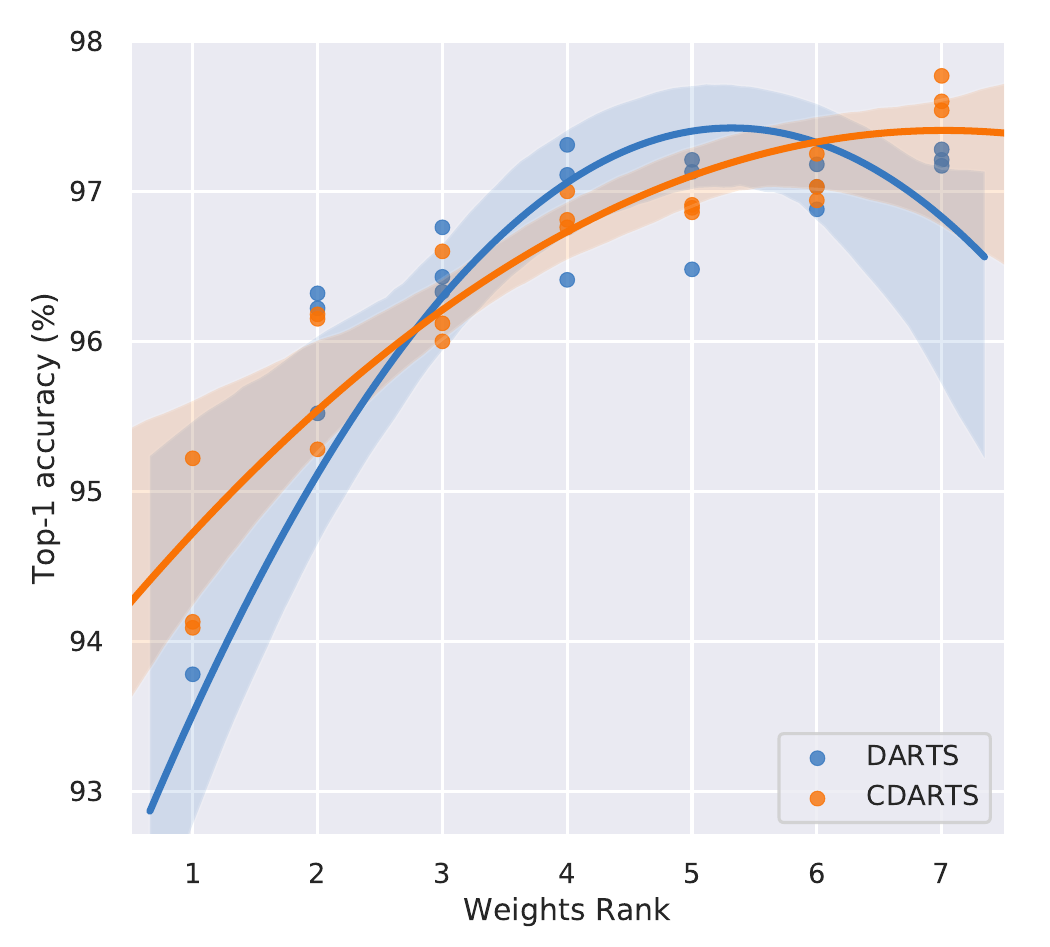}
			\vspace{-0.2cm} 
			\caption{Operation weights and retrain accuracy correlation analysis. Weights Rank indicates the weight sorting of the learned architecture from small to large.}
			\label{fig:corr}
		\end{minipage}
		\vspace{-0.cm}
		%\\[12pt]%设置两个表格之间的空白行距离
		\hspace{1em}
		\begin{minipage}{.45\textwidth}
			\renewcommand{\arraystretch}{1.3}
			\centering
			\vspace{0.2em}
			\includegraphics[height=0.68\textwidth]{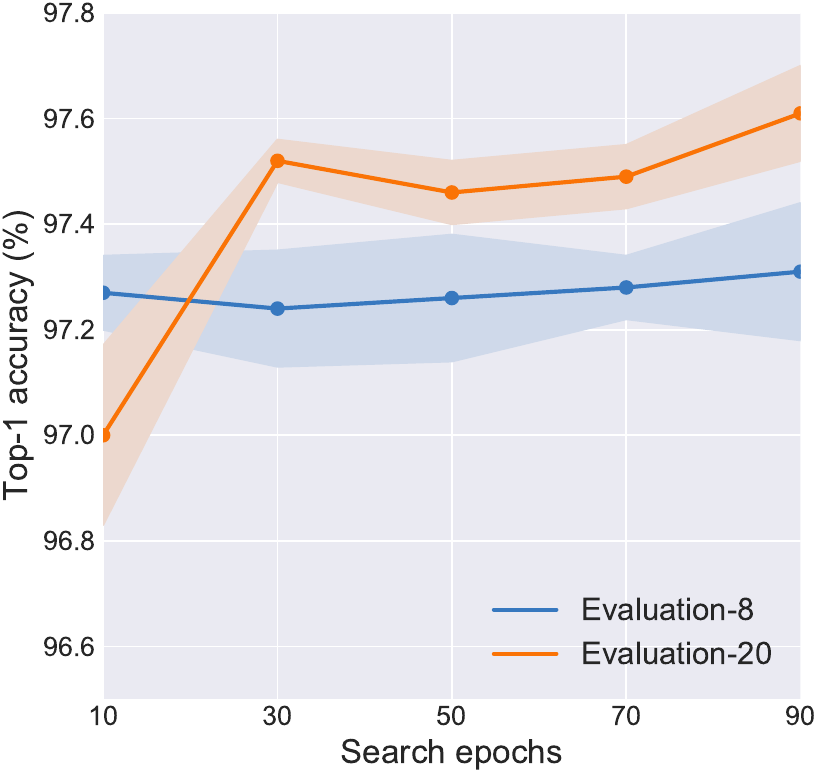}
			\vspace{-0.1em} 
			\caption{Ablation of the search epochs and the depth of evaluation network. Evaluation-8/20 represent the evaluation networks with 8/20 cells, respectively.}
			\label{fig:epoch}
		\end{minipage}
		\vspace{-1em}%设置第二个表格与正文下文的空白行距离
	\end{figure*}
	
	\begin{table}[t]
		\renewcommand{\arraystretch}{1.3}
		\centering
		\small
		\vspace{0.5em}
		\scalebox{0.8}
		{
			\begin{threeparttable}[b]
				% \resizebox{0.6\textwidth}{!}{
				\begin{tabular}{lccccccc}
					% \arrayrulecolor{blue}
					\toprule[1.2pt]
					\multirow{2}*{\textbf{Accuracy}} & \multicolumn{5}{c}{\textbf{Runs}} & \multirow{2}*{\textbf{ Mean$\pm$std}} \\
					\cline{2-6} & \textbf{\#1} & \textbf{\#2} & \textbf{\#3} & \textbf{\#4} & \textbf{\#5} \\
					\toprule[1.2pt]
					Top1(\%) &76.45 &75.90 &\textbf{76.61} &76.40 &75.94 & 76.26$\pm$0.29  \\
					Top5(\%) &92.83 &92.80 &\textbf{93.20} &93.03 &92.75 & 92.92$\pm$0.17 \\
					Params(M) &6.31 &5.99 &6.36 &6.32 &5.73 & 6.14$\pm$0.24 \\
					Flops(M) &718.68 &695.98 &725.20 &717.22 &651.45 & 701.70$\pm$26.97 \\
					
					\bottomrule[1.2pt]		
				\end{tabular}
			\end{threeparttable}
		}
		\caption{Evaluation of searched architectures in five independent search runs on ImageNet.}
		\label{table:imagruns}
	\end{table}

	% \begin{table}[t]
	% 	\renewcommand{\arraystretch}{1.3}
	% 	\centering
	% 	\small
	% 	\caption{Evaluation of searched architectures in five independent search runs on ImageNet.}
	% 	\label{table:imagruns}
	% 	\scalebox{0.9}
	% 	{
	% 		\begin{threeparttable}[b]
	% 			% \resizebox{0.6\textwidth}{!}{
	% 			\begin{tabular}{lccccccc}	
	% 				\toprule
	% 				\multirow{2}*{\textbf{Accuracy}} & \multicolumn{5}{c}{\textbf{Runs}} & \multirow{2}*{\textbf{ Mean$\pm$std}} \\
	% 				\cline{2-6} & \textbf{\#1} & \textbf{\#2} & \textbf{\#3} & \textbf{\#4} & \textbf{\#5} \\
	% 				\midrule
	% 				Top1(\%) &76.45 &75.90 &\textbf{76.61} &76.40 &75.94 & 76.26$\pm$0.29  \\
	% 				Top5(\%) &92.83 &92.80 &\textbf{93.20} &93.03 &92.75 & 92.92$\pm$0.17 \\
	% 				\bottomrule		
	% 			\end{tabular}
	% 		\end{threeparttable}
	% 	}
	% \end{table}

	The evaluation results of the searched architectures are reported in Tab.~\ref{tab:imagenet_results}. It shows that the architectures discoverd by our CDARTS is slightly better than the manually designed models, such as MobileNet-V2~\cite{sandler2018mobilenetv2} and ShuffleNet~\cite{zhang2018shufflenet}. Compared with automated methods, \emph{e.g.}, RL-based MobileNet-V3~\cite{mobilenetv3},
	our gradient-based CDARTS achieves comparable performance. MobileNet-V3 blends automatic search techniques with the interaction of human design, while CDARTS is purely automatic.
	%EfficientNets introduces human designs in the post-processing, but ours does not. 
	Moreover, CDARTS outperforms the prior DARTS~\cite{DARTS} by $3.0$ points in term of top-1 accuracy. This improvement is attributed to the proposed search algorithm. 
	%Compared with the recent PDARTS~\cite{PDARTS}, \textcolor{black}{our CDARTS is slightly superior, obtaining $0.7\%$ absolute improvement}. 
	It is observed that the \emph{flops} of CDARTS is a little higher than other DARTS methods. This is solely caused by the search algorithm itself, because the search space is the same among these DARTS methods. To follow the mobile setting~\cite{DARTS} comparison, \emph{i.e.}, the number of multiply-add operations in the model is restricted to be less than $600M$, we reduce the number of model channels from $48$ to $44$, while keeping other settings unchanged. We obtain $75.9\%$ top-1 accuracy with $571M$ \emph{flops}, which is still slightly superior than PDARTS~\cite{PDARTS} and PCDARTS~\cite{PCDARTS}. In addition, when training the discovered CIFAR10 cells on ImageNet, CDARTS obtains a top-1 accuracy of 75.6\%, 
	which is on par with ImageNet cell 76.3\%. This demonstrates the generalization potential of CDARTS.
	
	\vspace{-0.1cm}
	In Tab.~\ref{table:imagruns}, we present the results of five independent searches on ImageNet. We observe that the five runs of our method consistently exceed that of PDARTS (75.6\%) and PCDARTS (75.8\%). %Its  standard deviations of top-1 accuracy is 0.29\%. 
	The discovered cells on ImageNet are presented in Fig.~\ref{fig:imgcell}. 
	We observe that the cells discovered on ImageNet is deeper than the ones on CIFAR10, because the classification on ImageNet is more complex. This is aligned with the evidence that increasing network depth is beneficial for elevating model capability~\cite{he2016deep}. Moreover, the discovered cells on ImageNet contain larger convolution kernels (\emph{i.e.}, 5x5 sep\_conv), which is helpful to improve model capacity.

	\vspace{-0.3cm}
	\subsection{Ablation Study}
	\vspace{-0.2cm}
	
	\emph{Component-wise Analysis.} 
	To further understand the effects of the components in the proposed search algorithm, we test four variations of CDARTS on the ImageNet dataset. In particular, each variation corresponds to an optimization objective listed in Tab.~\ref{tab:loss}, and parameters for each model are tuned separately to obtain the optimal results.
	It is worth noting that the alternating optimization of $\mathcal{L}_{train}^S$ + $\mathcal{L}_{val}^S$ fail to search on ImageNet, whose cells are full of skip-connections. So we use the $\ell_1$-norm regularization to stabilize the searching process, and achieve a top-1 accuracy of 72.8\%. 
	By adding the proposed joint learning $\mathcal{L}_{val}^{S,E}$ into optimization, the performance is improved to 75.7\%. This indicates the effectiveness of the multi-level feature semantics guidance of the evaluation network, and the integration of the search and evaluation is beneficial to discover more robust architecture. Moreover, during joint training, updating the weights of the evaluation network can further improve the performance by 0.9\%.

	\emph{Correlation Analysis.} In differentiable architecture search methods, the operations and edges with weak attention (\emph{i.e.}, small weights) are considered as redundant and are pruned to obtain a compact architecture (\emph{i.e.}, top-$k$ discretization). However, it is not clear whether the operations and edges with weak attention are truly redundant, while strong attention indicates the high importance. Therefore, we conduct another experiment to evaluate the correlation between the architecture hyperparameter and the true performance of architectures. 
	Specifically, we sample a variety of cell architectures and rank them according to the learned weights of the architecture hyperparameter.
	The quantitative results shown in Fig.~\ref{fig:corr} demonstrate that our method obtains better correlation than DARTS~\cite{DARTS}. 
	We further compare the Kendall Rank Correlation Coefficient (\bm{$\tau$}) metric that evaluates the rank correlation of data pairs. We repeat the experiment six times with different seeds. The Kendall's \bm{$\tau$} of SPOS~\cite{spos} is 0.19~\cite{Cream}, while ours is 0.49, being 0.3 point superior to SPOS.
	To some extent, the architecture in CDARTS is able to reflect the relative ranking of architectures. But it is worth noting that CDARTS still cannot well distinguish the architectures that share close performance.

	\emph{Depth of Evaluation Network.}
	Due to the limitation of GPU memory, the search network of DARTS can only stack 8 cells, while the evaluation network contains 20 cells. This brings the so-called \emph{depth gap} issue studied in PDARTS~\cite{PDARTS}. 
	Such problem is not observed in our method, and we believe the reason is due to the integration of search and evaluation in a unified architecture.
	As shown in Fig.~\ref{fig:epoch}, we compare the performances of different numbers of cells in the evaluation network. It clearly shows that the 20-cell evaluation network (the red line) performs better than the 8-cell network (the green line). This indicates the proposed jointly training of the two networks mitigates the impacts of the depth gap.

	% \begin{figure}[t]
	% 	\centering
	% 	\includegraphics[width=0.4\textwidth]{figures/8_20.pdf}
	% 	\vspace{-0.cm} 
	% 	\caption{Ablation of search epochs and the depth of evaluation network. Evaluation-8/20 represent the evaluation networks with 8/20 cells, respectively.}
	% 	\label{fig:epoch}
	% \end{figure}
	
	\emph{Impact of Search Epoch.} In DARTS methods~\cite{DARTS,PCDARTS}, after certain searching epochs, the number of skip-connections increases
	dramatically in the selected architecture, which results in the search collapse~\cite{liang2019darts+}. We study the impact of the number of search epochs in our CDARTS. From Fig.~\ref{fig:epoch}, we can see that when the number of search epochs approaches 30, the performance becomes saturated, and the structure of the evaluation network tends to be stable. So the search epoch is set to 30 in experiments. 
	In addition, as the number of search epochs increases, CDARTS continues to search for better structures. Our method even achieves 97.77\% top-1 accuracy on CIFAR10 when the search epochs are 90.
	
	%\begin{figure*}[t]
	%	\centering
	%	\vspace{-0.cm}
	%	% \includegraphics[scale=0.95]{figures/arch.pdf}
	%	\includegraphics[width=0.8\textwidth]{figures/chain.pdf}
	%	\vspace{-0.cm} 
	%	\caption{(a) Previous NAS methods use pretrained models for knowledge distillation. (b) Our Introspective distillation enables knowledge transfer between architecture candidates.}
	%	\label{fig:chain}
	%	\vspace{-0.cm} 
	%\end{figure*}

	\subsection{Generality and Robustness}
	\subsubsection{Extension to Chain-structured Search Space}
	
	\begin{figure*}[t]
		\centering
		\vspace{-0.3cm}
		\includegraphics[height=0.35\textwidth]{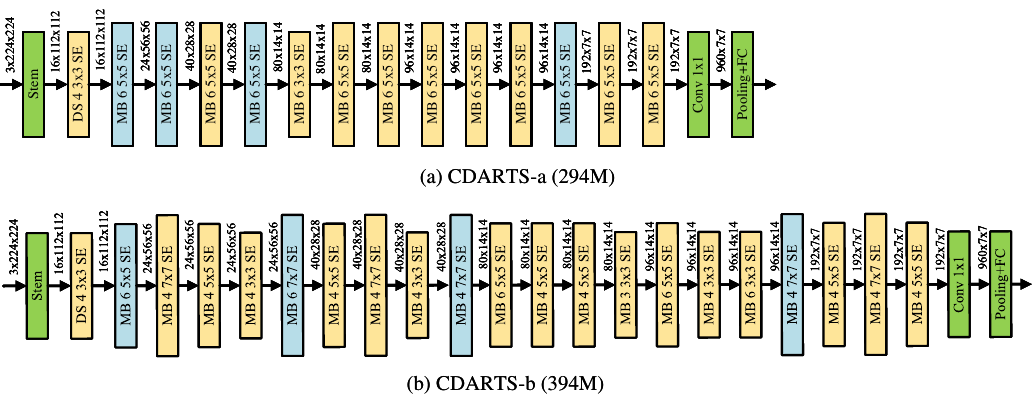}
		\vspace{-0.5em} 
		\caption{Discovered architectures. "$MB\ a\ b \times b$" represents the inverted bottleneck MBConv with the expand rate of a and kernel size of b. "$DS\ a\ b \times b$"  denotes the depthwise separable convolution with the expand rate of a and kernel size of b.}
		\label{fig:chain_arch}
		\vspace{-0.3cm} 
	\end{figure*}
	
	\begin{table}[t]
		% 	\vspace{-em}
		%\renewcommand{\arraystretch}{1.3}
		
		\vspace{-0.3cm}
		\small
		\centering
		\begin{tabular}{lcccc}
			% \arrayrulecolor{blue}
			% \begin{tabular}{l|llll}
			\toprule[1.2pt]
			$\mathcal{L}_{train}^S$ + $\mathcal{L}_{val}^S$ &\checkmark&\checkmark&\checkmark&\checkmark \\
			+ $\mathcal{L}_{Reg}$ & &\checkmark&\checkmark&\checkmark \\
			+ $\mathcal{L}_{val}^{S,E}$ &&&\checkmark&\checkmark \\
			+ $\mathcal{L}_{val}^{E}$ &&&&\checkmark \\
			\midrule
			Top-1 Acc. (\%) & - & 72.8 & 75.7 &76.6\\
			Params. (M) & - & 4.54 & 5.83 &6.36\\
			Flops (M) & - & 481.58 & 675.52 &725.20\\
			\bottomrule[1.2pt]
		\end{tabular}
		\caption{Ablation study on ImageNet.}
		\label{tab:loss}
		\vspace{-0.3cm}
	\end{table}
	
	A robust search algorithm should be capable of searching architectures over diverse search spaces. We migrate the introspective distillation searching method to a chain-structured search space~\cite{Survey}. The search space consisting of mobile inverted bottleneck MBConv~\cite{sandler2018mobilenetv2} and squeeze-excitation modules~\cite{hu2018senet}, which is the same as recent works~\cite{blockwise,ofa} for fair comparisons. There are seven basic operators, including MBConv with kernel sizes of \{3,5,7\} and expansion rates of \{4,6\}, and an additional skip connection to enable elastic depth of architectures. The space contains about $3.69 \times 10^{16}$ architecture candidates in total. For simplicity, we set SPOS~\cite{spos} as our baseline method and introduce a simple principle to select the teacher path.
	% The overall framework is visualized in Fig.~\ref{fig:chain}. 
	Specifically, the teacher path is first initialized with random paths within the search network. Then, for each batch, we randomly sample a single path from the search network and train the path under both the teacher path soft-label supervision and the ground truth supervision. After that, we evaluate the path on the validation dataset (a subset is used to save computation), and get its performance. If the current path performs superior to the teacher path and with less or the same \emph{Flops}, then the teacher network will be replaced by the current path. In this way, the final left teacher path is the optimal path of all sampled paths during the search.

	\begin{table}[t]
		\begin{center}
			\small
			\label{tab:imagenet_chain}
			%\hskip-0.3cm
			%\begin{tabular}{|p{4.3cm}|p{0.8cm}|p{0.8cm}|p{0.8cm}|}
			\begin{tabular}{l c c c c}
				% \arrayrulecolor{blue}
				\toprule[1.2pt]
				Model                                           &Params        & FLOPs     & Acc@1  &  Search   \\
				\toprule[1.2pt]
				 
				SPOS$^\dag$ \cite{spos}                      & 3.5M            & 319M      & 74.3\%  & 12  \\
				 
				DSNAS$^\dag$ \cite{hu2020dsnas}              & 3.5M             & 324M      & 74.4\%   & 17.5  \\
				 
				FBNet-C \cite{wu2019fbnet}                     & 5.5M             & 375M      & 74.9\% & 9   \\
				 
				MobileNetV3$^\sharp$ \cite{mobilenetv3}         & 5.3M          & 219M      & 75.2\%  & -   \\
				 
				MnasNet-A3$^\sharp$ \cite{tan2019mnasnet}               & 5.2M          & 403M      & 76.7\%  & -   \\
				 
				OFA$^\sharp$$^\dag$ \cite{ofa} & 5.8M          & 230M      & 76.9\%   & 55  \\
				 
				FairNAS-A$^\dag$ \cite{fairnas}                & 4.6M          & 388M      & 75.3\%  & 12   \\
				 
				MoGA-A$^\sharp$$^\dag$ \cite{moga}                      & 5.1M          & 304M      & 75.9\%  & 12   \\
				 
				PC-NAS-S$^\dag$ \cite{li2020improving}                & 5.1M          & -         & 76.8\%  & -   \\
				 
				MixNet-M$^\sharp$$^\dag$ \cite{tan2019mixconv}                 & 5.0M          & 360M      & 77.0\%  & -   \\
				 
				DNA-b$^\sharp$$^\dag$ \cite{blockwise}                        & 4.9M          & 406M      & 77.5\%  & 24.6  \\
				EfficientNet-B0$^\sharp$ \cite{tan2019efficientnet}     & 5.3M          & 399M      & 77.1\%  & -   \\
				ProxylessNAS$^\dag$ \cite{cai2018proxylessnas}        & 7.1M          & 465M      & 75.1\% & 15    \\
				 
				SCARLET-A$^\dag$ \cite{chu2019scarletnas}             & 6.7M          & 365M      & 76.9\%  & 10  \\
				% AutoFormer-tiny \cite{chen2021autoformer}             & 5.7M          & 1.3G      & 74.7\%  & 32  \\
				\hline
				\textbf{CDARTS-a$^\sharp$$^\dag$} & 7.0M          & 294M      & 77.4\%  & 12  \\
				\textbf{CDARTS-b$^\sharp$$^\dag$} & 6.4M          & 394M      & \textbf{78.2\%} & 12  \\
				\bottomrule[1.2pt]
			\end{tabular}
		\end{center}
		\vspace{-0.5em}
		\caption{Comparisons of chain-structured search space models on ImageNet. The input size is $224\times 224$. $\sharp$: using the SE module. The unit of search time is GPU day. $\dag$: the search space contains kernel-size-7 operation.}
		\label{tab:imagenet_chain}
		\vspace{-0.85em}
	\end{table}
	
	%  It is worth noting that the search space for different search methods is not exactly the same.
	We train the search network for 120 epochs using the following settings: the SGD optimizer~\cite{SGD} with momentum 0.9 and weight decay 4e-5, initial learning rate 0.5 with linear annealing. The discovered architectures are presented in Fig.~\ref{fig:chain_arch} and we retrain them for 350 epochs on ImageNet using similar settings as EfficientNet~\cite{tan2019efficientnet}: RMSProp optimizer with momentum 0.9 and decay 0.9, weight decay 1e-5, dropout ratio 0.2, initial learning rate 0.04 with a warmup~\cite{goyal2017accurate} in the first 3 epochs and a cosine annealing. We also adopt the AutoAugment~\cite{cubuk2018autoaugment} and exponential moving average techniques. We use 8 Nvidia Tesla V100 GPUs with a batch size of 1,024 for the retraining.

	It is worth noting that there are a few recent works that leverage knowledge distillation techniques to boost searching~\cite{ofa,blockwise}. As shown in Tab~\ref{tab:imagenet_chain}, our introspective distillation shows superior performance over these methods. Specifically, DNA-b~\cite{blockwise} recruits EfficientNet-B7~\cite{tan2019efficientnet}, a very high-performance third-party model, as the teacher and achieves 77.5\% top-1 accuracy, while our method
	(CDARTS-b) gets a superior accuracy of 78.2\% without using any pre-trained model.
	% Compared with recent transformer method, CDARTS-b is 3.5\% higher than AutoFormer-tiny \cite{chen2021autoformer} with less Flops.
	Our model also surpasses EfficientNet-B0 by 1.1\% with nearly the same model size. The leading performance indicates that the proposed introspective distillation method also works well on the chain-structured search space.

	\vspace{-0.2cm}
	\subsubsection{Extension to Big Model}
	To further unleash the power of the searched architectures, we enlarge the evaluation network channels and train from scratch with more data augmentations~\cite{cubuk2018autoaugment,zhang2018mixup}, which is denoted as \emph{big} model.
	
	\setlength{\tabcolsep}{4pt}
	\begin{table*}[!t]
		\begin{center}
			\small
			\caption{Object detection results of various drop-in backbones on the COCO val2017. Acc represents the top-1 accuracy on ImageNet. $\dagger$ reported by \cite{fairnas}}
			\label{tab:coco_retina}
			\scalebox{1.1}{
				\begin{tabular}{*{1}{l}c*{9}{c}}
					% \arrayrulecolor{blue}
					% \hline\noalign{\smallskip}
					\toprule[1.2pt]
					Backbones & Input Size & FLOPS(G)  & Params(M)    & mAP(\%) & AP$_{50}$ & AP$_{75}$ & AP$_S$ & AP$_M$ & AP$_L$ &Top-1(\%)  \\
					%\begin{scriptsize}
					\toprule[1.2pt]
					% \hline\noalign{\smallskip}
					MobileNetV2$^{\dagger}$ \cite{sandler2018mobilenetv2} & 1280$\times$800 & 6.1 & 3.4 & 28.3 & 46.7 & 29.3 & 14.8 & 30.7 & 38.1 & 72.0\\
					 
					SPOS$^{\dagger}$ \cite{spos} & 1280$\times$800 & 7.4 & 4.3  & 30.7 & 49.8 & 32.2 & 15.4 &33.9 & 41.6& 75.0\\
					%			FBNet-C \cite{wu2018fbnet} &375 &5.5 & 74.9\\
					 
					MobileNetV3$^{\dagger}$ \cite{mobilenetv3} & 1280$\times$800 & 4.5 & - & 29.9& 49.3 & 30.8 & 14.9 & 33.3 & 41.1  & 75.2\\
					 
					MnasNet-A2$^{\dagger}$ \cite{tan2019mnasnet} & 1280$\times$800 & 6.9 & 4.8  & 30.5 & 50.2 & 32.0 & 16.6 & 34.1 & 41.1 & 75.6\\
					%			EfficientNetB0 \cite{tan2019efficientnet} & 390 & 5.3&76.3& \\
					 
					MixNet-M$^{\dagger}$ \cite{tan2019mixconv} & 1280$\times$800 & 7.3 & 5.0  & 31.3& 51.7 & 32.4& 17.0 & 35.0 & 41.9 & 77.0  \\
					 
					FairNAS-A$^{\dagger}$ \cite{fairnas} & 1280$\times$800 & 8.0 & 5.9  & 32.4 & 52.4 & 33.9 & 17.2 & 36.3 & 43.2 & 77.5\\
					 
					MixPath-A \cite{chu2020mixpath} & 1280$\times$800 & 7.1 & 5.0  &31.5& 51.3 & 33.2 & 17.4& 35.3& 41.8 & 76.9 \\
					% \cdashline{1-7}[0.8pt/2pt]
					\hline
					\textbf{CDARTS-a}  & 1280$\times$800 &  6.0 & 7.0  &35.2& 55.5 & 37.5 & 19.8& 38.7& 47.5 & 77.4 \\
					\textbf{CDARTS-b}  & 1280$\times$800 & 8.1 & 6.4  &\textbf{36.2}& \textbf{56.7} & \textbf{38.3} & \textbf{20.9}& \textbf{39.8}& \textbf{48.5} & \textbf{78.2} \\
					\bottomrule[1.2pt]
				\end{tabular}
			}
		\end{center}
	\end{table*}
	\setlength{\tabcolsep}{1.4pt}
	
	\begin{table}[!t]
		\renewcommand{\arraystretch}{1.3}
		\vspace{-0.2cm}
		\caption{Top-1 accuracy of big models.}
		\label{tab:big_model}
		\small
		\centering
		\begin{tabular}{lccc}
			%\hline
			\toprule[1.2pt]
			%\multirow{2}{*}{Method} & \multicolumn{2}{c}{ \small Top-1 Acc. (\%)} \\
			%\cmidrule(lr){2-3} &  \scalebox{0.9}{CIFAR10}  & \scalebox{0.9}{ImageNet}\\
			Method &  CIFAR10  & ImageNet \\
			%\hline
			\toprule[1.2pt]
			DARTS~\cite{DARTS}  & 97.95  & -- \\
			PDARTS~\cite{PDARTS}  & 98.00  & 80.04 \\
			PCDARTS~\cite{PCDARTS}  & 97.92  & 79.81 \\
			\midrule
			\textbf{CDARTS}  & \textbf{98.32}  & \textbf{81.12} \\
			%\hline
			\bottomrule[1.2pt]
		\end{tabular}
	    \vspace{-0.2cm}
	\end{table}

	\textit{Big Model on CIFAR10}
	% we increase the width of the evaluation network. 
	We enlarge the numbers of feature channels from $36$ to $50$. After $2000$ epochs training, the CDARTS-BIG model  obtains impressive performance improvements. On CIFAR10, the test accuracy increases from $97.52$\% to $98.32$\%. This improvement further verifies the effectiveness of the proposed method. In addition, to have a fair comparison, we retrain the evaluation networks discovered by other DARTS methods, such as DARTS, PDARTS and PCDARTS with the same \emph{big} setting. As presented in Tab.~\ref{tab:big_model}, our CDARTS performs the best among them.

	\textit{Big Model on ImageNet}
	% the number of cells from $14$ to $20$, 
	We increase the channel number from 48 to 96 and the input image size from 224$\times$224 to 320$\times$320 to construct a big model with $5.6B$ \emph{flops}.
	This big model is trained for 250 epochs, and obtains a $4.86$ points absolute improvement over the original CDARTS, achieving $81.12$ top-1 accuracy on ImageNet, which is comparable to EfficientNet-B2~\cite{tan2019efficientnet}. %This verifies the effectiveness and generalizability of the proposed search method. 
	Moreover, following the same settings, we train the big models of PDARTS and PCDARTS. 
	Their performances are inferior to our CDARTS, as presented in Tab.~\ref{tab:big_model}. 
	It should be noted that the original DARTS cells have been proved in many experiments to be inferior to the improved version, e.g. PDARTS \cite{PDARTS}, PCDARTS \cite{PCDARTS}. In order to save computation resources, we did not train big model of DARTS on ImageNet.
	These results confirm again the generalization potential and effectiveness of CDARTS.

	\begin{table}[!t]
	\vspace{-0.2cm}
	\centering
	\caption{Comparisons of models for semantic segmentation on the Cityscapes validation set.}
	\scalebox{0.92}{
		\begin{tabular}{c  c  c  c  c }
			% \arrayrulecolor{blue}
			\toprule[1.5pt]
			\multirow{2}{*}{Methods} & \multirow{2}{*}{Encoder} & Flops  & Params & mIoU \\
			& & {(G)}  & {(M)} & (\%) \\
			\toprule[1.5pt]
			ESPNetV2~\cite{mehta2018espnet}& ESPNetV2 & 2.7 & 1.3 & 66.2  \\
			ICNet~\cite{zhao2018icnet}& PSPNet50 & - & - & 69.5  \\
			% BiSeNet~\cite{yu2018bisenet}& 768x1536 & - & - & 69.0  \\
			% Fast-SCNN~\cite{poudel2019fast}& 1024x2048 & - & - & 68.6  \\
			% DF1-Seg-d8~\cite{li2019partial}& 1024x2048  & - & - & 72.4   \\
			HRNet~\cite{WangSCJDZLMTWLX19}& HRNetV2 & 31.1 & 1.5 & 70.3 \\
			%$\text{HRNetV2}_{W40}$~\cite{WangSCJDZLMTWLX19}& 1024x2048  & 493.2 & 45.2 & 80.2 \\
			CAS~\cite{zhang2019customizable}& CAS & - & - & 71.6\\
			MobilenetV3~\cite{howard2019searching}& Large & 9.7 & 1.5  & 72.4 \\
			 
			FasterSeg~\cite{chen2020fasterseg}& FasterSeg & 28.2 & 4.4 & 73.1 \\
			BiSeNetV2-L~\cite{yu2021bisenet}& - & 21.2 & - & 73.4  \\
			SqueezeNAS~\cite{shaw2019squeezenas}& LAT Large & 19.6 & 1.9  & 73.6 \\
			SwiftNetRN-18~\cite{orsic2019defense} &  ResNet18 & 104.0 & 11.8 & 75.4  \\
			SegFormer~\cite{xie2021segformer}& MiT-B0 & 125.5 & 3.8 & 76.2 \\
			\midrule
			\textbf{CDARTS(Ours)} & CDARTS-b & 20.7 & 5.9 & \textbf{78.1} \\
			\bottomrule[1.5pt]
		\end{tabular}\label{tab:cityscapes}
	}
	\vspace{-0.3cm}
	\end{table}

	\subsubsection{Extension to Object Detection}
	We also transfer the proposed method to the downstream object detection task, which is a fundamental task in the vision community. We use the discovered architecture and corresponding weights pre-trained on ImageNet as a drop-in replacement for the backbone feature extractor in RetinaNet~\cite{lin2017focal}. The capabilities of various light-weight backbones are compared on the COCO benchmark~\cite{coco}. We fine-tune the searched model on the \emph{train2017} set (118k images) and evaluate on the \emph{val2017} set (5k images) with the batch size of 16 on 8 V100 GPUs. Similar to~\cite{fairnas}, the schedule is the default 1$\times$, \textit{i.e.}, 12 epochs. The initial learning rate is set to 0.02, and then decayed with the scaling factor of 0.1 at the 8-th and 11-th epoch. The optimizer is SGD with momentum 0.9 and weight decay 1e-4. We use the MMDetection toolbox~\cite{chen2019mmdetection} based on PyTorch~\cite{paszke2019pytorch}.
	
	Results are summarized in Tab.~\ref{tab:coco_retina}. Specifically, CDARTS-a surpasses MobileNetV2 by 6.9\% while using fewer Flops. Compared to MnasNet~\cite{tan2019mnasnet}, CDARTS-a utilizes 19\% fewer Flops yet achieves 4.7\% higher performance. CDARTS-b even achieves 36.2\% mAP on the COCO val2017, which surpasses other methods by a large margin. With similar flops and parameters, CDARTS-b is 3.8\% higher than FairNAS-A~\cite{fairnas}, suggesting the architecture has good generalization ability.

	\subsubsection{Extension to Semantic Segmentation}
	To evaluate the generalization ability of the architectures found by CDARTS, we transfer the architectures to the downstream semantic segmentation~\cite{li2019expectation,li2020improving,li2020spatial} task. We leverage the discovered architecture and corresponding weights pre-trained on ImageNet as a drop-in replacement for the encoder in FasterSeg~\cite{chen2020fasterseg}. The mean intersection over union per class (mIoU) is used as the metric for semantic segmentation.

	\textbf{Cityscapes}~\cite{Cordts2016Cityscapes} is a popular dataset which contains a diverse set of stereo video sequences recorded in street scenes from 50 different cities. It has 5,000 high quality pixel-level annotations. There are 2,975 images for training, 500 images for validation. And for testing, it offers 1,525 images without ground-truth for a fair comparison. The dense annotation contains 19 classes for each pixel.
	We evaluate our CDARTS on Cityscapes validation set with the original image resolution of $1024\times2048$. In Table ~\ref{tab:cityscapes}, we see the superior mIoU and model size of our CDARTS. Without any inference trick, our CDARTS achieves 78.1\% mIoU, which is 5.0\% better than FasterSeg~\cite{chen2020fasterseg}. Specifically, our CDARTS surpasses SegFormer~\cite{xie2021segformer} by 1.9\% with much fewer Flops.

	\textbf{ADE20K}~\cite{zhou2017scene} has 20k images for training, 2k images for validation and 3k images for testing. It is used in ImageNet scene parsing challenge 2016, and has 150 classes and diverse scenes with 1,038 image-level.
	We also evaluate our CDARTS on the ADE20K validation set with the original image resolution of $640\times640$. From Table~\ref{tab:ade20k}, we can see that our model is much smaller than others. Specifically, our CDARTS achieves 40.4\% mIoU with 5.9G Flops, which is 3.0\% better than SegFormer~\cite{xie2021segformer}.

	\begin{table}[!t]
		\vspace{-0.2cm}
		\centering
		\caption{Comparisons of models for semantic segmentation on the ADE20K validation set.}
		\scalebox{0.92}{
			\begin{tabular}{c  c  c  c  c }
				% \arrayrulecolor{blue}
				\toprule[1.5pt]
				\multirow{2}{*}{Methods} & \multirow{2}{*}{Encoder} & Flops  & Params & mIoU \\
				& & {(G)}  & {(M)} & (\%) \\
				\toprule[1.5pt]
				FCN~\cite{long2015fully}& MobileNetV2 & 39.6 & 9.8 & 19.7 \\
				PSPNet~\cite{zhao2017pyramid}& MobileNetV2 & 52.9 & 13.7 & 29.6 \\
				DeepLabV3+~\cite{chen2018encoder}& MobileNetV2 & 69.4 & 15.4 & 34.0 \\
				SegFormer~\cite{xie2021segformer}& MiT-B0 & 8.4 & 3.8 & 37.4 \\
				\midrule
				\textbf{CDARTS(Ours)} & CDARTS-b & 5.9 & 2.7 & \textbf{40.4} \\
				\bottomrule[1.5pt]
			\end{tabular}\label{tab:ade20k}
		}
		\vspace{-0.5cm}
	\end{table}
	
	\vspace{-0.4cm}
	\subsection{Discussion}
	\vspace{-0.2cm}
	\textit{Computation cost.} 
	The computation cost of CDARTS is comparable to DARTSV1. Both of them adopt the first-order optimization method~\cite{DARTS} to train the networks. Compared with the original DARTS~\cite{DARTS}, we have an additional evaluation network and update it along with the search network in the search process. We denote the computation cost of updating the search network as $\mathcal{C}(|W_S|)$, which is the same as DARTSV1. In the search network cell, the number of edges between nodes is $ED_S$ and the number of operations in each edge is $OP_S$. The number of stacked cells in the search network is $N_S$. %Correspondingly, these of the evaluation network are $ED_E$, $OP_E$ and $N_E$. 
	Correspondingly, these factors of the evaluation network are denoted as $ED_E$, $OP_E$ and $N_E$.
	Then the cost of the evaluation network $\mathcal{C}(|W_E|)$ is:
	\begin{equation} \label{eq:complexity}
	\centering
	\mathcal{C}(|W_E|) = \frac{N_E \cdot ED_E \cdot OP_E}{N_S \cdot ED_S \cdot OP_S} \cdot \mathcal{C}(|W_S|)
	\vspace{-0.cm}
	\end{equation}
	In the experiments on CIFAR10, the $ED_E$, $OP_E$ and $N_E$ are set to 8, 1 and 20, respectively, while the $ED_S$, $OP_S$ and $N_S$ are set to 14, 8 and 8, respectively. Hence, the cost of the evaluation network $\mathcal{C}(|W_E|)$ is about $\frac{1}{6}$ of $\mathcal{C}(|W_S|)$. Compared to these two networks, the cost of the embedding modules is much smaller and can be ignored in cost estimation. Considering the initialization stage of the evaluation network (\emph{i.e.}, one training epoch for CIFAR10), the final cost of the evaluation network is about $\frac{1}{3}$ of $\mathcal{C}(|W_S|)$. In ImageNet, the final cost of the evaluation network is about $\frac{1}{2}$ of $\mathcal{C}(|W_S|)$. Therefore, compared to DARTSV1~\cite{DARTS}, the complexity of CDARTS is only increased by $\sim$0.3 times.  
	
	% \HL{(It is very weird to say something like ``$\frac{1}{3}$ of $\mathcal{O}(|W_S|)$", since the big O notation should ignore the constant coefficient (ie, $\frac{1}{3}$ in this case).)}
	
	\textit{Fast search speed on ImageNet.}
	It is worth noting that our method takes about five hours to complete the search with eight GPUs on ImageNet. Such a fast search speed mainly attributes to the following three reasons. First, we use the fast first-order optimization algorithm~\cite{DARTS} when updating the hyperparameter of architecture. Second, following PCDARTS~\cite{PCDARTS}, only 10\% data in each category are used. Besides, the evaluation network is relatively lightweight and adopts the weight sharing strategy to speed up network training, so the extra computational cost of updating its parameters is small.

	\vspace{-0.4cm}
	\section{Conclusions}
	\vspace{-0.1cm}
	\label{sec:ccl}
	In this work, motivated by the separation problem of the search and evaluation networks in DARTS, we have proposed a cyclic differentiable architecture search algorithm that integrates the two networks into a unified architecture. The alternating joint learning enables the search of  architectures to fit the final evaluation network. Experiments on three different search spaces demonstrate the efficacy of the proposed algorithm and searched architectures, which achieve competitive performance on CIFAR, ImageNet and NATS-Bench. In future work, we will consider adding more constraints on prioritized path selection, such as both Params and latency, thus improving the flexibility and user-friendliness of the search method.

	\textbf{Acknowledgments.} This work was jointly supported by National Key Research and Development Program of China Grant No. 2018AAA0100400, National Natural Science Foundation of China (61721004, U1803261, and 61976132), Beijing Nova Program (Z201100006820079), Key Research Program of Frontier Sciences CAS Grant No. ZDBS-LY-JSC032, and CAS-AIR.

	\bibliographystyle{IEEEtran}
	\bibliography{egbib}

\begin{thebibliography}{100}
\providecommand{\url}[1]{#1}
\csname url@samestyle\endcsname
\providecommand{\newblock}{\relax}
\providecommand{\bibinfo}[2]{#2}
\providecommand{\BIBentrySTDinterwordspacing}{\spaceskip=0pt\relax}
\providecommand{\BIBentryALTinterwordstretchfactor}{4}
\providecommand{\BIBentryALTinterwordspacing}{\spaceskip=\fontdimen2\font plus
\BIBentryALTinterwordstretchfactor\fontdimen3\font minus
  \fontdimen4\font\relax}
\providecommand{\BIBforeignlanguage}[2]{{%
\expandafter\ifx\csname l@#1\endcsname\relax
\typeout{** WARNING: IEEEtran.bst: No hyphenation pattern has been}%
\typeout{** loaded for the language `#1'. Using the pattern for}%
\typeout{** the default language instead.}%
\else
\language=\csname l@#1\endcsname
\fi
#2}}
\providecommand{\BIBdecl}{\relax}
\BIBdecl

\bibitem{NASNet}
B.~Zoph, V.~Vasudevan, J.~Shlens, and Q.~V. Le, ``Learning transferable
  architectures for scalable image recognition,'' in \emph{\textit{CVPR}},
  2018, pp. 8697--8710.

\bibitem{chang2020data}
J.~Chang, Y.~Guo, M.~Gaofeng, Z.~Lin, S.~XIANG, C.~Pan \emph{et~al.}, ``Data:
  Differentiable architecture approximation with distribution guided
  sampling,'' \emph{\textit{IEEE Transactions on Pattern Analysis and Machine
  Intelligence}}, 2020.

\bibitem{ghiasi2019fpn}
G.~Ghiasi, T.-Y. Lin, and Q.~V. Le, ``Nas-fpn: Learning scalable feature
  pyramid architecture for object detection,'' in \emph{\textit{CVPR}}, 2019,
  pp. 7036--7045.

\bibitem{chen2019detnas}
Y.~Chen, T.~Yang, X.~Zhang, G.~Meng, X.~Xiao, and J.~Sun, ``Detnas: Backbone
  search for object detection,'' vol.~32, 2019, pp. 6642--6652.

\bibitem{liu2019auto}
C.~Liu, L.-C. Chen, F.~Schroff, H.~Adam, W.~Hua, A.~L. Yuille, and L.~Fei-Fei,
  ``Auto-deeplab: Hierarchical neural architecture search for semantic image
  segmentation,'' in \emph{\textit{CVPR}}, 2019, pp. 82--92.

\bibitem{nekrasov2019fast}
V.~Nekrasov, H.~Chen, C.~Shen, and I.~Reid, ``Fast neural architecture search
  of compact semantic segmentation models via auxiliary cells,'' in
  \emph{\textit{CVPR}}, 2019, pp. 9126--9135.

\bibitem{DARTS}
H.~Liu, K.~Simonyan, and Y.~Yang, ``{DARTS}: Differentiable architecture
  search,'' in \emph{\textit{ICLR}}, 2019.

\bibitem{ren2021comprehensive}
P.~Ren, Y.~Xiao, X.~Chang, P.-Y. Huang, Z.~Li, X.~Chen, and X.~Wang, ``A
  comprehensive survey of neural architecture search: Challenges and
  solutions,'' \emph{ACM Computing Surveys (CSUR)}, vol.~54, no.~4, pp. 1--34,
  2021.

\bibitem{NAS_RL}
B.~Zoph and Q.~V. Le, ``Neural architecture search with reinforcement
  learning,'' in \emph{\textit{ICML}}, 2017.

\bibitem{PNAS}
C.~Liu, B.~Zoph, M.~Neumann, J.~Shlens, W.~Hua, L.-J. Li, L.~Fei-Fei,
  A.~Yuille, J.~Huang, and K.~Murphy, ``Progressive neural architecture
  search,'' in \emph{\textit{ECCV}}, 2018, pp. 19--34.

\bibitem{kandasamy2018neural}
K.~Kandasamy, W.~Neiswanger, J.~Schneider, B.~Poczos, and E.~P. Xing, ``Neural
  architecture search with bayesian optimisation and optimal transport,'' in
  \emph{\textit{NeurIPS}}, 2018.

\bibitem{Survey}
T.~Elsken, J.~H. Metzen, and F.~Hutter, ``Neural architecture search: A
  survey.'' \emph{\textit{Journal of Machine Learning Research}}, vol.~20,
  no.~55, pp. 1--21, 2019.

\bibitem{Survey2}
M.~Wistuba, A.~Rawat, and T.~Pedapati, ``A survey on neural architecture
  search,'' \emph{\textit{arXiv:1905.01392}}, 2019.

\bibitem{cai2018proxylessnas}
H.~Cai, L.~Zhu, and S.~Han, ``Proxyless{NAS}: Direct neural architecture search
  on target task and hardware,'' in \emph{\textit{ICLR}}, 2019.

\bibitem{PDARTS}
X.~Chen, L.~Xie, J.~Wu, and Q.~Tian, ``Progressive differentiable architecture
  search: Bridging the depth gap between search and evaluation,'' in
  \emph{\textit{ICCV}}, 2019, pp. 1294--1303.

\bibitem{yang2021towards}
Y.~Yang, S.~You, H.~Li, F.~Wang, C.~Qian, and Z.~Lin, ``Towards improving the
  consistency, efficiency, and flexibility of differentiable neural
  architecture search,'' in \emph{\textit{CVPR}}, 2021, pp. 6667--6676.

\bibitem{xie2018snas}
S.~Xie, H.~Zheng, C.~Liu, and L.~Lin, ``{SNAS}: stochastic neural architecture
  search,'' in \emph{\textit{ICLR}}, 2019.

\bibitem{GDAS}
X.~Dong and Y.~Yang, ``Searching for a robust neural architecture in four gpu
  hours,'' in \emph{\textit{CVPR}}, 2019, pp. 1761--1770.

\bibitem{AutoHAS}
X.~Dong, M.~Tan, A.~W. Yu, D.~Peng, B.~Gabrys, and Q.~V. Le, ``Autohas:
  Efficient hyperparameter and architecture search,'' in \emph{\textit{ICLR}},
  2021.

\bibitem{randomNAS}
L.~Li and A.~Talwalkar, ``Random search and reproducibility for neural
  architecture search,'' in \emph{\textit{UAI}}.\hskip 1em plus 0.5em minus
  0.4em\relax PMLR, 2020, pp. 367--377.

\bibitem{Understanding}
A.~Zela, T.~Elsken, T.~Saikia, Y.~Marrakchi, T.~Brox, and F.~Hutter,
  ``Understanding and robustifying differentiable architecture search,'' in
  \emph{\textit{ICLR}}, 2020.

\bibitem{liang2019darts+}
H.~Liang, S.~Zhang, J.~Sun, X.~He, W.~Huang, K.~Zhuang, and Z.~Li, ``Darts+:
  Improved differentiable architecture search with early stopping,''
  \emph{\textit{arXiv:1909.06035}}, 2019.

\bibitem{wang2021rethinking}
R.~Wang, M.~Cheng, X.~Chen, X.~Tang, and C.-J. Hsieh, ``Rethinking architecture
  selection in differentiable {NAS},'' in \emph{\textit{ICLR}}, 2021.

\bibitem{RandomSPOS}
C.~Sciuto, K.~Yu, M.~Jaggi, C.~Musat, and M.~Salzmann, ``Evaluating the search
  phase of neural architecture search,'' in \emph{\textit{ICLR}}, 2020.

\bibitem{hard}
A.~Yang, P.~M. Esperança, and F.~M. Carlucci, ``Nas evaluation is
  frustratingly hard,'' in \emph{\textit{ICLR}}, 2020.

\bibitem{xie2020weight}
L.~Xie, X.~Chen, K.~Bi, L.~Wei, Y.~Xu, Z.~Chen, L.~Wang, A.~Xiao, J.~Chang,
  X.~Zhang \emph{et~al.}, ``Weight-sharing neural architecture search: A battle
  to shrink the optimization gap,'' \emph{arXiv preprint arXiv:2008.01475},
  2020.

\bibitem{liu2019search}
Y.~Liu, X.~Jia, M.~Tan, R.~Vemulapalli, Y.~Zhu, B.~Green, and X.~Wang, ``Search
  to distill: Pearls are everywhere but not the eyes,'' in
  \emph{\textit{CVPR}}, 2020, pp. 7539--7548.

\bibitem{blockwise}
C.~Li, J.~Peng, L.~Yuan, G.~Wang, X.~Liang, L.~Lin, and X.~Chang,
  ``Block-wisely supervised neural architecture search with knowledge
  distillation,'' in \emph{\textit{CVPR}}, 2020, pp. 1989--1998.

\bibitem{krizhevsky2009learning}
A.~Krizhevsky, G.~Hinton \emph{et~al.}, ``Learning multiple layers of features
  from tiny images,'' Citeseer, Tech. Rep., 2009.

\bibitem{russakovsky2015imagenet}
O.~Russakovsky, J.~Deng, H.~Su, J.~Krause, S.~Satheesh, S.~Ma, Z.~Huang,
  A.~Karpathy, A.~Khosla, M.~Bernstein \emph{et~al.}, ``{ImageNet} large scale
  visual recognition challenge,'' \emph{\textit{International Journal of
  Computer Vision}}, vol. 115, no.~3, pp. 211--252, 2015.

\bibitem{bench201}
X.~Dong and Y.~Yang, ``Nas-bench-201: Extending the scope of reproducible
  neural architecture search,'' in \emph{\textit{ICLR}}, 2020.

\bibitem{mobilenetv3}
A.~Howard, M.~Sandler, G.~Chu, L.-C. Chen, B.~Chen, M.~Tan, W.~Wang, Y.~Zhu,
  R.~Pang, V.~Vasudevan \emph{et~al.}, ``Searching for mobilenetv3,'' in
  \emph{\textit{ICCV}}, 2019, pp. 1314--1324.

\bibitem{SETN}
X.~Dong and Y.~Yang, ``One-shot neural architecture search via self-evaluated
  template network,'' in \emph{\textit{ICCV}, pages={3681--3690}}, 2019.

\bibitem{REINFORCE}
R.~J. Williams, ``Simple statistical gradient-following algorithms for
  connectionist reinforcement learning,'' \emph{\textit{Machine learning}},
  vol.~8, no. 3-4, pp. 229--256, 1992.

\bibitem{ENAS}
H.~Pham, M.~Guan, B.~Zoph, Q.~Le, and J.~Dean, ``Efficient neural architecture
  search via parameters sharing,'' in \emph{\textit{ICLR}}.\hskip 1em plus
  0.5em minus 0.4em\relax PMLR, 2018, pp. 4095--4104.

\bibitem{tan2019efficientnet}
M.~Tan and Q.~Le, ``Efficientnet: Rethinking model scaling for convolutional
  neural networks,'' in \emph{\textit{ICLR}}.\hskip 1em plus 0.5em minus
  0.4em\relax PMLR, 2019, pp. 6105--6114.

\bibitem{xie2021segformer}
E.~Xie, W.~Wang, Z.~Yu, A.~Anandkumar, J.~M. Alvarez, and P.~Luo, ``Segformer:
  Simple and efficient design for semantic segmentation with transformers,'' in
  \emph{\textit{NeurIPS}}, 2021.

\bibitem{chu2019fairdarts}
X.~Chu, T.~Zhou, B.~Zhang, and J.~Li, ``Fair darts: Eliminating unfair
  advantages in differentiable architecture search,'' in
  \emph{\textit{ECCV}}.\hskip 1em plus 0.5em minus 0.4em\relax Springer, 2020,
  pp. 465--480.

\bibitem{hinton2015distilling}
G.~Hinton, O.~Vinyals, and J.~Dean, ``Distilling the knowledge in a neural
  network,'' \emph{\textit{arXiv:1503.02531}}, 2015.

\bibitem{shen2015multi}
L.~Shen, G.~Sun, Q.~Huang, S.~Wang, Z.~Lin, and E.~Wu, ``Multi-level
  discriminative dictionary learning with application to large scale image
  classification,'' \emph{\textit{IEEE Transactions on Image Processing}},
  vol.~24, no.~10, pp. 3109--3123, 2015.

\bibitem{yang2020dynamical}
Y.~Yang, J.~Wu, H.~Li, X.~Li, T.~Shen, and Z.~Lin, ``Dynamical system inspired
  adaptive time stepping controller for residual network families,'' in
  \emph{\textit{AAAI}}, vol.~34, no.~04, 2020, pp. 6648--6655.

\bibitem{liu2020deep}
L.~Liu, W.~Ouyang, X.~Wang, P.~Fieguth, J.~Chen, X.~Liu, and
  M.~Pietik{\"a}inen, ``Deep learning for generic object detection: A survey,''
  \emph{\textit{International Journal of Computer Vision}}, vol. 128, no.~2,
  pp. 261--318, 2020.

\bibitem{ren2015faster}
S.~{Ren}, K.~{He}, R.~{Girshick}, and J.~{Sun}, ``Faster r-cnn: Towards
  real-time object detection with region proposal networks,''
  \emph{\textit{IEEE Transactions on Pattern Analysis and Machine
  Intelligence}}, vol.~39, no.~6, pp. 1137--1149, 2017.

\bibitem{liu2010robust}
G.~Liu, Z.~Lin, Y.~Yu \emph{et~al.}, ``Robust subspace segmentation by low-rank
  representation.'' in \emph{\textit{ICML}}, vol.~1.\hskip 1em plus 0.5em minus
  0.4em\relax Citeseer, 2010, p.~8.

\bibitem{yang2020sognet}
Y.~Yang, H.~Li, X.~Li, Q.~Zhao, J.~Wu, and Z.~Lin, ``Sognet: Scene overlap
  graph network for panoptic segmentation,'' in \emph{\textit{AAAI}}, vol.~34,
  no.~07, 2020, pp. 12\,637--12\,644.

\bibitem{li2019expectation}
X.~Li, Z.~Zhong, J.~Wu, Y.~Yang, Z.~Lin, and H.~Liu, ``Expectation-maximization
  attention networks for semantic segmentation,'' in \emph{\textit{ICCV}},
  2019, pp. 9167--9176.

\bibitem{tracking_pami}
T.~{Wang} and H.~{Ling}, ``Gracker: A graph-based planar object tracker,''
  \emph{IEEE Transactions on Pattern Analysis and Machine Intelligence},
  vol.~40, no.~6, pp. 1494--1501, 2018.

\bibitem{lin_pami}
W.~{Wang}, J.~{Shen}, and H.~{Ling}, ``A deep network solution for attention
  and aesthetics aware photo cropping,'' \emph{\textit{IEEE Transactions on
  Pattern Analysis and Machine Intelligence}}, vol.~41, no.~7, pp. 1531--1544,
  2019.

\bibitem{li2021towards}
X.~Li, X.~Li, A.~You, L.~Zhang, G.~Cheng, K.~Yang, Y.~Tong, and Z.~Lin,
  ``Towards efficient scene understanding via squeeze reasoning,''
  \emph{\textit{IEEE Transactions on Image Processing}}, vol.~30, pp.
  7050--7063, 2021.

\bibitem{zhong2018joint}
Z.~Zhong, T.~Shen, Y.~Yang, Z.~Lin, and C.~Zhang, ``Joint sub-bands learning
  with clique structures for wavelet domain super-resolution,'' vol.~31, 2018.

\bibitem{han2015learning}
S.~Han, J.~Pool, J.~Tran, and W.~Dally, ``Learning both weights and connections
  for efficient neural network,'' in \emph{\textit{NeurIPS}}, 2015.

\bibitem{adanet}
C.~Cortes, X.~Gonzalvo, V.~Kuznetsov, M.~Mohri, and S.~Yang, ``Adanet: Adaptive
  structural learning of artificial neural networks,'' in
  \emph{\textit{ICLR}}.\hskip 1em plus 0.5em minus 0.4em\relax PMLR, 2017, pp.
  874--883.

\bibitem{smithson2016neural}
S.~C. Smithson, G.~Yang, W.~J. Gross, and B.~H. Meyer, ``Neural networks
  designing neural networks: multi-objective hyper-parameter optimization,'' in
  \emph{\textit{Proceedings of the 35th International Conference on
  Computer-Aided Design}}, 2016, pp. 1--8.

\bibitem{saxena2016convolutional}
S.~Saxena and J.~Verbeek, ``Convolutional neural fabrics,'' in
  \emph{\textit{NeurIPS}}, vol.~29, 2016, pp. 4053--4061.

\bibitem{baker2016designing}
B.~Baker, O.~Gupta, N.~Naik, and R.~Raskar, ``Designing neural network
  architectures using reinforcement learning,'' in \emph{\textit{ICLR}}, 2017.

\bibitem{xie2017genetic}
L.~Xie and A.~Yuille, ``Genetic cnn,'' in \emph{\textit{ICCV}}, pp. 1379--1388.

\bibitem{tan2019mnasnet}
M.~Tan, B.~Chen, R.~Pang, V.~Vasudevan, M.~Sandler, A.~Howard, and Q.~V. Le,
  ``Mnasnet: Platform-aware neural architecture search for mobile,'' in
  \emph{\textit{CVPR}}, 2019, pp. 2820--2828.

\bibitem{bignas}
J.~Yu, P.~Jin, H.~Liu, G.~Bender, P.-J. Kindermans, M.~Tan, T.~Huang, X.~Song,
  R.~Pang, and Q.~Le, ``Bignas: Scaling up neural architecture search with big
  single-stage models,'' in \emph{\textit{ECCV}}.\hskip 1em plus 0.5em minus
  0.4em\relax Springer, 2020, pp. 702--717.

\bibitem{dai2020fbnetv3}
X.~Dai, A.~Wan, P.~Zhang, B.~Wu, Z.~He, Z.~Wei, K.~Chen, Y.~Tian, M.~Yu,
  P.~Vajda \emph{et~al.}, ``Fbnetv3: Joint architecture-recipe search using
  neural acquisition function,'' \emph{\textit{arXiv preprint
  arXiv:2006.02049}}, 2020.

\bibitem{sandler2018mobilenetv2}
M.~Sandler, A.~Howard, M.~Zhu, A.~Zhmoginov, and L.-C. Chen, ``Mobilenetv2:
  Inverted residuals and linear bottlenecks,'' in \emph{\textit{CVPR}}, 2018,
  pp. 4510--4520.

\bibitem{ofa}
H.~Cai, C.~Gan, T.~Wang, Z.~Zhang, and S.~Han, ``Once for all: Train one
  network and specialize it for efficient deployment,'' in
  \emph{\textit{ICLR}}, 2020.

\bibitem{he2016deep}
K.~He, X.~Zhang, S.~Ren, and J.~Sun, ``Deep residual learning for image
  recognition,'' in \emph{\textit{CVPR}}, 2016, pp. 770--778.

\bibitem{huang2017densely}
G.~Huang, Z.~Liu, L.~Van Der~Maaten, and K.~Q. Weinberger, ``Densely connected
  convolutional networks,'' in \emph{\textit{CVPR}}, 2017, pp. 4700--4708.

\bibitem{hu2018senet}
J.~Hu, L.~Shen, and G.~Sun, ``Squeeze-and-excitation networks,'' in
  \emph{\textit{CVPR}}, 2018, pp. 7132--7141.

\bibitem{zhong2018practical}
Z.~Zhong, J.~Yan, W.~Wu, J.~Shao, and C.-L. Liu, ``Practical block-wise neural
  network architecture generation,'' in \emph{\textit{CVPR}}, 2018, pp.
  2423--2432.

\bibitem{real2019regularized}
E.~Real, A.~Aggarwal, Y.~Huang, and Q.~V. Le, ``Regularized evolution for image
  classifier architecture search,'' in \emph{\textit{AAAI}}, vol.~33, no.~01,
  2019, pp. 4780--4789.

\bibitem{nasbench101}
C.~Ying, A.~Klein, E.~Christiansen, E.~Real, K.~Murphy, and F.~Hutter,
  ``Nas-bench-101: Towards reproducible neural architecture search,'' in
  \emph{\textit{ICLR}}.\hskip 1em plus 0.5em minus 0.4em\relax PMLR, 2019, pp.
  7105--7114.

\bibitem{nasbench1shot1}
A.~Zela, J.~Siems, and F.~Hutter, ``Nas-bench-1shot1: Benchmarking and
  dissecting one-shot neural architecture search,'' in \emph{\textit{ICLR}},
  2020.

\bibitem{wang2018evolving}
B.~Wang, Y.~Sun, B.~Xue, and M.~Zhang, ``Evolving deep convolutional neural
  networks by variable-length particle swarm optimization for image
  classification,'' in \emph{\textit{IEEE Congress on Evolutionary Computation
  (CEC) }}.\hskip 1em plus 0.5em minus 0.4em\relax IEEE, 2018, pp. 1--8.

\bibitem{liang2018evolutionary}
J.~Liang, E.~Meyerson, and R.~Miikkulainen, ``Evolutionary architecture search
  for deep multitask networks,'' in \emph{\textit{Proceedings of the Genetic
  and Evolutionary Computation Conference}}, 2018, pp. 466--473.

\bibitem{sun2018evolving}
Y.~Sun, G.~G. Yen, and Z.~Yi, ``Evolving unsupervised deep neural networks for
  learning meaningful representations,'' \emph{\textit{IEEE Transactions on
  Evolutionary Computation}}, vol.~23, no.~1, pp. 89--103, 2018.

\bibitem{dong2018dpp}
J.-D. Dong, A.-C. Cheng, D.-C. Juan, W.~Wei, and M.~Sun, ``Dpp-net:
  Device-aware progressive search for pareto-optimal neural architectures,'' in
  \emph{\textit{ECCV}}, 2018, pp. 517--531.

\bibitem{before_pnas}
H.~Liu, K.~Simonyan, O.~Vinyals, C.~Fernando, and K.~Kavukcuoglu,
  ``Hierarchical representations for efficient architecture search,'' in
  \emph{\textit{ICLR}}, 2018.

\bibitem{SMASH}
A.~Brock, T.~Lim, J.~M. Ritchie, and N.~Weston, ``Smash: one-shot model
  architecture search through hypernetworks,'' in \emph{\textit{ICLR}}, 2018.

\bibitem{bender2018understanding}
G.~Bender, P.-J. Kindermans, B.~Zoph, V.~Vasudevan, and Q.~Le, ``Understanding
  and simplifying one-shot architecture search,'' in
  \emph{\textit{ICLR}}.\hskip 1em plus 0.5em minus 0.4em\relax PMLR, 2018, pp.
  550--559.

\bibitem{wu2019fbnet}
B.~Wu, X.~Dai, P.~Zhang, Y.~Wang, F.~Sun, Y.~Wu, Y.~Tian, P.~Vajda, Y.~Jia, and
  K.~Keutzer, ``Fbnet: Hardware-aware efficient convnet design via
  differentiable neural architecture search,'' in \emph{\textit{CVPR}}, 2019,
  pp. 10\,734--10\,742.

\bibitem{spos}
Z.~Guo, X.~Zhang, H.~Mu, W.~Heng, Z.~Liu, Y.~Wei, and J.~Sun, ``Single path
  one-shot neural architecture search with uniform sampling,'' in
  \emph{\textit{ECCV}}.\hskip 1em plus 0.5em minus 0.4em\relax Springer, 2020,
  pp. 544--560.

\bibitem{moga}
X.~Chu, B.~Zhang, and R.~Xu, ``Moga: Searching beyond mobilenetv3,'' in
  \emph{\textit{ICASSP}}.\hskip 1em plus 0.5em minus 0.4em\relax IEEE, 2020,
  pp. 4042--4046.

\bibitem{fairnas}
X.~Chu, B.~Zhang, R.~Xu, and J.~Li, ``Fairnas: Rethinking evaluation fairness
  of weight sharing neural architecture search,'' in \emph{\textit{ICCV}},
  2019, pp. 12\,239--12\,248.

\bibitem{li2020improving}
X.~Li, C.~Lin, C.~Li, M.~Sun, W.~Wu, J.~Yan, and W.~Ouyang, ``Improving
  one-shot nas by suppressing the posterior fading,'' in \emph{\textit{CVPR}},
  2020, pp. 13\,836--13\,845.

\bibitem{PCDARTS}
Y.~Xu, L.~Xie, X.~Zhang, X.~Chen, G.-J. Qi, Q.~Tian, and H.~Xiong,
  ``{PC-DARTS}: Partial channel connections for memory-efficient architecture
  search,'' in \emph{\textit{ICLR}}, 2020.

\bibitem{yang2020ista}
Y.~Yang, H.~Li, S.~You, F.~Wang, C.~Qian, and Z.~Lin, ``Ista-nas: Efficient and
  consistent neural architecture search by sparse coding,'' in
  \emph{\textit{NeurIPS}}, vol.~33, 2020.

\bibitem{istrate2019tapas}
R.~Istrate, F.~Scheidegger, G.~Mariani, D.~Nikolopoulos, C.~Bekas, and A.~C.~I.
  Malossi, ``Tapas: Train-less accuracy predictor for architecture search,'' in
  \emph{\textit{AAAI}}, vol.~33, no.~01, 2019, pp. 3927--3934.

\bibitem{wong2018transfer}
C.~Wong, N.~Houlsby, Y.~Lu, and A.~Gesmundo, ``Transfer learning with neural
  automl,'' in \emph{\textit{NeurIPS}}, 2018.

\bibitem{wistuba2019xfernas}
M.~Wistuba, ``Xfernas: Transfer neural architecture search,'' in \emph{Joint
  European Conference on Machine Learning and Knowledge Discovery in
  Databases}.\hskip 1em plus 0.5em minus 0.4em\relax Springer, 2020, pp.
  247--262.

\bibitem{peng2019efficient}
J.~Peng, M.~Sun, Z.~Zhang, T.~Tan, and J.~Yan, ``Efficient neural architecture
  transformation searchin channel-level for object detection,'' in
  \emph{\textit{NeurIPS}}, 2020.

\bibitem{fang2020fna++}
J.~Fang, Y.~Sun, Q.~Zhang, K.~Peng, Y.~Li, W.~Liu, and X.~Wang, ``Fna++: Fast
  network adaptation via parameter remapping and architecture search,''
  \emph{\textit{IEEE Transactions on Pattern Analysis and Machine
  Intelligence}}, 2020.

\bibitem{lu2021neural}
Z.~Lu, G.~Sreekumar, E.~Goodman, W.~Banzhaf, K.~Deb, and V.~N. Boddeti,
  ``Neural architecture transfer,'' \emph{\textit{IEEE Transactions on Pattern
  Analysis and Machine Intelligence}}, 2021.

\bibitem{Cream}
H.~Peng, H.~Du, H.~Yu, Q.~Li, J.~Liao, and J.~Fu, ``Cream of the crop:
  Distilling prioritized paths for one-shot neural architecture search,''
  \emph{NeurIPS}, 2020.

\bibitem{meta_pseudo}
H.~Pham, Z.~Dai, Q.~Xie, and Q.~V. Le, ``Meta pseudo labels,'' in
  \emph{\textit{CVPR}}, 2021, pp. 11\,557--11\,568.

\bibitem{sciuto2019evaluating}
C.~Sciuto, K.~Yu, M.~Jaggi, C.~Musat, and M.~Salzmann, ``Evaluating the search
  phase of neural architecture search,'' in \emph{\textit{ICLR}}, 2019.

\bibitem{krizhevsky2012imagenet}
A.~Krizhevsky, I.~Sutskever, and G.~E. Hinton, ``Imagenet classification with
  deep convolutional neural networks,'' in \emph{\textit{NeurIPS}}, vol.~25,
  2012, pp. 1097--1105.

\bibitem{kingma2014adam}
D.~P. Kingma and J.~Ba, ``Adam: A method for stochastic optimization,'' in
  \emph{\textit{ICLR}}, 2015.

\bibitem{paszke2019pytorch}
A.~Paszke, S.~Gross, F.~Massa, A.~Lerer, J.~Bradbury, G.~Chanan, T.~Killeen,
  Z.~Lin, N.~Gimelshein, L.~Antiga \emph{et~al.}, ``Pytorch: An imperative
  style, high-performance deep learning library,'' in \emph{\textit{NeurIPS}},
  vol.~32, 2019, pp. 8026--8037.

\bibitem{dong2021nats}
X.~Dong, L.~Liu, K.~Musial, and B.~Gabrys, ``Nats-bench: Benchmarking nas
  algorithms for architecture topology and size,'' \emph{\textit{IEEE
  transactions on pattern analysis and machine intelligence}}, 2021.

\bibitem{RS}
J.~Bergstra and Y.~Bengio, ``Random search for hyper-parameter optimization,''
  \emph{\textit{Journal of machine learning research}}, vol.~13, no. Feb, pp.
  281--305, 2012.

\bibitem{BOHB}
S.~Falkner, A.~Klein, and F.~Hutter, ``Bohb: Robust and efficient
  hyperparameter optimization at scale,'' in \emph{\textit{ICLR}}.\hskip 1em
  plus 0.5em minus 0.4em\relax PMLR, 2018, pp. 1437--1446.

\bibitem{chrabaszcz2017downsampled}
P.~Chrabaszcz, I.~Loshchilov, and F.~Hutter, ``A downsampled variant of
  imagenet as an alternative to the cifar datasets,''
  \emph{\textit{arXiv:1707.08819}}, 2017.

\bibitem{zagoruyko2016wide}
\BIBentryALTinterwordspacing
S.~Zagoruyko and N.~Komodakis, ``Wide residual networks,'' in \emph{Proceedings
  of the British Machine Vision Conference (BMVC)}, E.~R.~H. Richard C.~Wilson
  and W.~A.~P. Smith, Eds.\hskip 1em plus 0.5em minus 0.4em\relax BMVA Press,
  September 2016, pp. 87.1--87.12. [Online]. Available:
  \url{https://dx.doi.org/10.5244/C.30.87}
\BIBentrySTDinterwordspacing

\bibitem{xie2017aggregated}
S.~Xie, R.~Girshick, P.~Doll{\'a}r, Z.~Tu, and K.~He, ``Aggregated residual
  transformations for deep neural networks,'' in \emph{\textit{CVPR}}, 2017,
  pp. 1492--1500.

\bibitem{luo2018neural}
R.~Luo, F.~Tian, T.~Qin, E.~Chen, and T.-Y. Liu, ``Neural architecture
  optimization,'' in \emph{\textit{NeurIPS}}, 2018.

\bibitem{xu2020latency}
Y.~Xu, L.~Xie, X.~Zhang, X.~Chen, B.~Shi, Q.~Tian, and H.~Xiong,
  ``Latency-aware differentiable neural architecture search,'' \emph{arXiv
  preprint arXiv:2001.06392}, 2020.

\bibitem{loshchilov2016sgdr}
I.~Loshchilov and F.~Hutter, ``Sgdr: Stochastic gradient descent with warm
  restarts,'' in \emph{\textit{ICLR}}, 2017.

\bibitem{SGD}
H.~Robbins and S.~Monro, ``A stochastic approximation method,''
  \emph{\textit{The annals of mathematical statistics}}, pp. 400--407, 1951.

\bibitem{larsson2016fractalnet}
G.~Larsson, M.~Maire, and G.~Shakhnarovich, ``Fractalnet: Ultra-deep neural
  networks without residuals,'' 2017.

\bibitem{simonyan2014very}
K.~Simonyan and A.~Zisserman, ``Very deep convolutional networks for
  large-scale image recognition,'' in \emph{\textit{ICLR}}, 2015.

\bibitem{devries2017improved}
T.~DeVries and G.~W. Taylor, ``Improved regularization of convolutional neural
  networks with cutout,'' \emph{\textit{arXiv:1708.04552}}, 2017.

\bibitem{szegedy2015going}
C.~Szegedy, W.~Liu, Y.~Jia, P.~Sermanet, S.~Reed, D.~Anguelov, D.~Erhan,
  V.~Vanhoucke, and A.~Rabinovich, ``Going deeper with convolutions,'' in
  \emph{\textit{CVPR}}, 2015, pp. 1--9.

\bibitem{gholami2018squeezenext}
A.~Gholami, K.~Kwon, B.~Wu, Z.~Tai, X.~Yue, P.~Jin, S.~Zhao, and K.~Keutzer,
  ``Squeezenext: Hardware-aware neural network design,'' in \emph{\textit{CVPR
  Workshops}}, 2018, pp. 1638--1647.

\bibitem{shufflenetv2}
N.~Ma, X.~Zhang, H.-T. Zheng, and J.~Sun, ``Shufflenet v2: Practical guidelines
  for efficient cnn architecture design,'' in \emph{\textit{ECCV},
  pages={116--131}}, 2018.

\bibitem{nayman2019xnas}
N.~Nayman, A.~Noy, T.~Ridnik, I.~Friedman, R.~Jin, and L.~Zelnik, ``Xnas:
  Neural architecture search with expert advice,'' in \emph{\textit{NeurIPS}},
  2019.

\bibitem{noy2019asap}
A.~Noy, N.~Nayman, T.~Ridnik, N.~Zamir, S.~Doveh, I.~Friedman, R.~Giryes, and
  L.~Zelnik-Manor, ``Asap: Architecture search, anneal and prune,''
  \emph{\textit{arXiv:1904.04123}}, 2019.

\bibitem{howard2017mobilenets}
A.~G. Howard, M.~Zhu, B.~Chen, D.~Kalenichenko, W.~Wang, T.~Weyand,
  M.~Andreetto, and H.~Adam, ``Mobilenets: Efficient convolutional neural
  networks for mobile vision applications,'' \emph{\textit{arXiv:1704.04861}},
  2017.

\bibitem{goyal2017accurate}
P.~Goyal, P.~Doll{\'a}r, R.~Girshick, P.~Noordhuis, L.~Wesolowski, A.~Kyrola,
  A.~Tulloch, Y.~Jia, and K.~He, ``Accurate, large minibatch sgd: Training
  imagenet in 1 hour,'' \emph{\textit{arXiv:1706.02677}}, 2017.

\bibitem{szegedy2016rethinking}
C.~Szegedy, V.~Vanhoucke, S.~Ioffe, J.~Shlens, and Z.~Wojna, ``Rethinking the
  inception architecture for computer vision,'' in \emph{\textit{CVPR}}, 2016,
  pp. 2818--2826.

\bibitem{zhang2018shufflenet}
X.~Zhang, X.~Zhou, M.~Lin, and J.~Sun, ``Shufflenet: An extremely efficient
  convolutional neural network for mobile devices,'' in \emph{\textit{CVPR}},
  2018, pp. 6848--6856.

\bibitem{hu2020dsnas}
S.~Hu, S.~Xie, H.~Zheng, C.~Liu, J.~Shi, X.~Liu, and D.~Lin, ``Dsnas: Direct
  neural architecture search without parameter retraining,'' in
  \emph{\textit{CVPR}}, 2020, pp. 12\,084--12\,092.

\bibitem{tan2019mixconv}
M.~Tan and Q.~V. Le, ``Mixconv: Mixed depthwise convolutional kernels,'' in
  \emph{\textit{BMVC}}, 2019.

\bibitem{chu2019scarletnas}
X.~Chu, B.~Zhang, Q.~Li, R.~Xu, and X.~Li, ``Scarlet-nas: bridging the gap
  between stability and scalability in weight-sharing neural architecture
  search,'' in \emph{\textit{ICCV}, pages={317--325}}, 2021.

\bibitem{cubuk2018autoaugment}
E.~D. Cubuk, B.~Zoph, D.~Mane, V.~Vasudevan, and Q.~V. Le, ``Autoaugment:
  Learning augmentation strategies from data,'' in \emph{\textit{CVPR}}, 2019,
  pp. 113--123.

\bibitem{zhang2018mixup}
H.~Zhang, M.~Cisse, Y.~N. Dauphin, and D.~Lopez-Paz, ``mixup: Beyond empirical
  risk minimization,'' in \emph{\textit{ICLR}}, 2018.

\bibitem{chu2020mixpath}
X.~Chu, X.~Li, Y.~Lu, B.~Zhang, and J.~Li, ``Mixpath: A unified approach for
  one-shot neural architecture search,'' \emph{\textit{arXiv:2001.05887}},
  2020.

\bibitem{mehta2018espnet}
S.~Mehta, M.~Rastegari, A.~Caspi, L.~Shapiro, and H.~Hajishirzi, ``Espnet:
  Efficient spatial pyramid of dilated convolutions for semantic
  segmentation,'' in \emph{\textit{ECCV}}, 2018, pp. 552--568.

\bibitem{zhao2018icnet}
H.~Zhao, X.~Qi, X.~Shen, J.~Shi, and J.~Jia, ``Icnet for real-time semantic
  segmentation on high-resolution images,'' in \emph{\textit{ECCV}}, 2018, pp.
  405--420.

\bibitem{WangSCJDZLMTWLX19}
J.~Wang, K.~Sun, T.~Cheng, B.~Jiang, C.~Deng, Y.~Zhao, D.~Liu, Y.~Mu, M.~Tan,
  X.~Wang, W.~Liu, and B.~Xiao, ``Deep high-resolution representation learning
  for visual recognition,'' \emph{\textit{TPAMI}}, 2019.

\bibitem{zhang2019customizable}
Y.~Zhang, Z.~Qiu, J.~Liu, T.~Yao, D.~Liu, and T.~Mei, ``Customizable
  architecture search for semantic segmentation,'' in \emph{\textit{CVPR}},
  2019, pp. 11\,641--11\,650.

\bibitem{howard2019searching}
A.~Howard, M.~Sandler, G.~Chu, L.-C. Chen, B.~Chen, M.~Tan, W.~Wang, Y.~Zhu,
  R.~Pang, V.~Vasudevan \emph{et~al.}, ``Searching for mobilenetv3,'' in
  \emph{\textit{ICCV}}, 2019, pp. 1314--1324.

\bibitem{chen2020fasterseg}
W.~Chen, X.~Gong, X.~Liu, Q.~Zhang, Y.~Li, and Z.~Wang, ``Fasterseg: Searching
  for faster real-time semantic segmentation,'' in \emph{\textit{ICLR}}, 2020.

\bibitem{yu2021bisenet}
C.~Yu, C.~Gao, J.~Wang, G.~Yu, C.~Shen, and N.~Sang, ``Bisenet v2: Bilateral
  network with guided aggregation for real-time semantic segmentation,''
  \emph{\textit{International Journal of Computer Vision}}, pp. 1--18, 2021.

\bibitem{shaw2019squeezenas}
A.~Shaw, D.~Hunter, F.~Landola, and S.~Sidhu, ``Squeezenas: Fast neural
  architecture search for faster semantic segmentation,'' in \emph{\textit{ICCV
  Workshops}}, 2019, pp. 0--0.

\bibitem{orsic2019defense}
M.~Orsic, I.~Kreso, P.~Bevandic, and S.~Segvic, ``In defense of pre-trained
  imagenet architectures for real-time semantic segmentation of road-driving
  images,'' in \emph{\textit{CVPR}}, 2019, pp. 12\,607--12\,616.

\bibitem{lin2017focal}
T.-Y. Lin, P.~Goyal, R.~Girshick, K.~He, and P.~Doll{\'a}r, ``Focal loss for
  dense object detection,'' in \emph{\textit{ICCV}, pages={2980--2988}}, 2017.

\bibitem{coco}
T.-Y. Lin, M.~Maire, S.~Belongie, J.~Hays, P.~Perona, D.~Ramanan,
  P.~Doll{\'a}r, and C.~L. Zitnick, ``Microsoft coco: Common objects in
  context,'' in \emph{\textit{ECCV}}.\hskip 1em plus 0.5em minus 0.4em\relax
  Springer, 2014, pp. 740--755.

\bibitem{chen2019mmdetection}
K.~Chen, J.~Wang, J.~Pang, Y.~Cao, Y.~Xiong, X.~Li, S.~Sun, W.~Feng, Z.~Liu,
  J.~Xu \emph{et~al.}, ``Mmdetection: Open mmlab detection toolbox and
  benchmark,'' \emph{arXiv preprint arXiv:1906.07155}, 2019.

\bibitem{li2020spatial}
X.~Li, Y.~Yang, Q.~Zhao, T.~Shen, Z.~Lin, and H.~Liu, ``Spatial pyramid based
  graph reasoning for semantic segmentation,'' in \emph{\textit{CVPR}}, 2020,
  pp. 8950--8959.

\bibitem{Cordts2016Cityscapes}
M.~Cordts, M.~Omran, S.~Ramos, T.~Rehfeld, M.~Enzweiler, R.~Benenson,
  U.~Franke, S.~Roth, and B.~Schiele, ``The cityscapes dataset for semantic
  urban scene understanding,'' in \emph{\textit{CVPR}}, 2016, pp. 3213--3223.

\bibitem{zhou2017scene}
B.~Zhou, H.~Zhao, X.~Puig, S.~Fidler, A.~Barriuso, and A.~Torralba, ``Scene
  parsing through ade20k dataset,'' in \emph{\textit{CVPR}}, 2017, pp.
  633--641.

\bibitem{long2015fully}
J.~Long, E.~Shelhamer, and T.~Darrell, ``Fully convolutional networks for
  semantic segmentation,'' in \emph{\textit{CVPR}}, 2015, pp. 3431--3440.

\bibitem{zhao2017pyramid}
H.~Zhao, J.~Shi, X.~Qi, X.~Wang, and J.~Jia, ``Pyramid scene parsing network,''
  in \emph{\textit{CVPR}}, 2017, pp. 2881--2890.

\bibitem{chen2018encoder}
L.-C. Chen, Y.~Zhu, G.~Papandreou, F.~Schroff, and H.~Adam, ``Encoder-decoder
  with atrous separable convolution for semantic image segmentation,'' in
  \emph{\textit{ECCV}}, 2018, pp. 801--818.

\end{thebibliography}

	\vspace{-1.8cm}
	\begin{IEEEbiography}[{\includegraphics[width=1in,height=1.25in,clip,keepaspectratio]{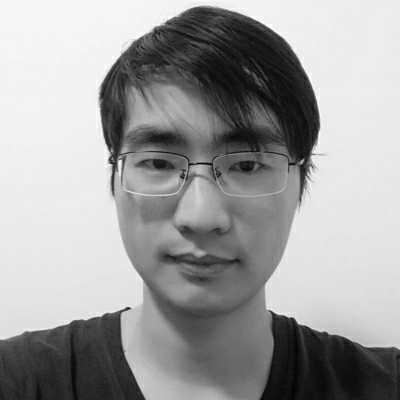}}]{Hongyuan Yu} received the BSc degree from the Nankai University (NKU), Tianjin, China, in 2017. He is currently pursuing the Ph.D. degree with the Institute of Automation, Chinese Academy of Sciences (CASIA), Beijing, China, and the University of Chinese Academy of Sciences (UCAS), Beijing, China. His research interests include machine learning and pattern recognition. 
	\end{IEEEbiography}
	% He received the Best Paper Runner-up Award from the International Conference on Document Analysis and Recognition (ICDAR) 2019.
	
	\vspace{-1.6cm}
	\begin{IEEEbiography}[{\includegraphics[width=1in,height=1.25in,clip,keepaspectratio]{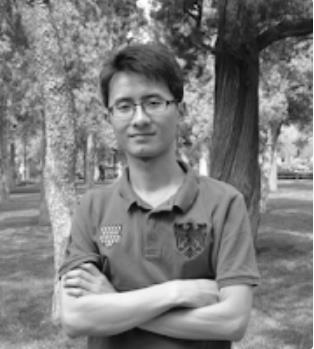}}]{Houwen Peng}
		is a senior researcher working on computer vision and deep learning at Microsoft Research as of 2018. Before that he was a senior engineer at Qualcomm AI Research. He received Ph.D. from NLPR, Institution of Automation, Chinese Academy of Sciences in 2016. From 2015 to 2016, he worked as a visiting research scholar at Temple University. His research interests include tiny and efficient deep learning, video object tracking, segmentation and detection, vision transformer, neural architecture search, model compression, vision-language intelligence, saliency detection, etc. He has served as the area chairs /senior program committee member for ACM Multimedia and AAAI.
	\end{IEEEbiography}
	
	\vspace{-1.1cm}
	\begin{IEEEbiography}[{\includegraphics[width=1in,height=1.25in,clip,keepaspectratio]{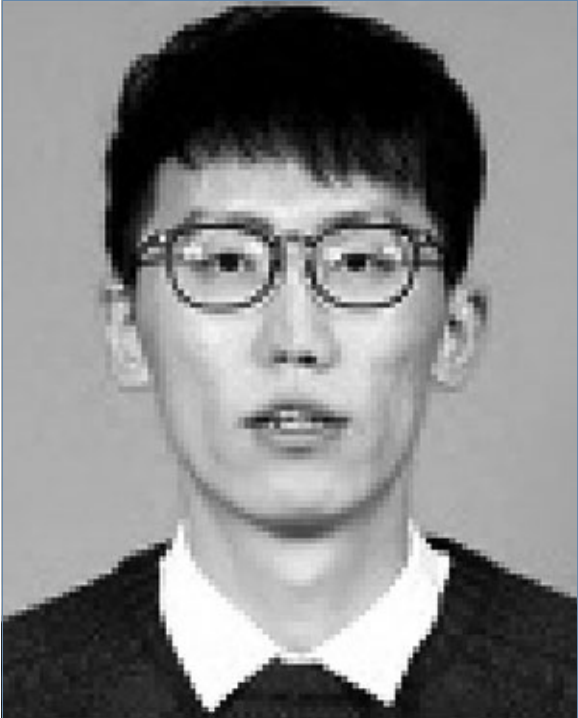}}]{Yan Huang} received the PhD degree from the University of Chinese Academy of Sciences (UCAS), in 2017.  Since July 2017, He has joined the National Laboratory of Pattern Recognition (NLPR), Institute of Automation, Chinese Academy of Sciences (CASIA) as an assistant professor. His research interests include machine learning and pattern recognition. He has published papers in the leading international journals and conferences such as the IEEE Transactions on Pattern Analysis and Machine Intelligence, the IEEE Transactions on Image Processing, NeurIPS, CVPR, ICCV and ECCV.
	%\textbf{	BSc degree from the University of Electronic Science and Technology of China (UESTC), in 2012, and the PhD degree from the University of Chinese Academy of Sciences (UCAS), in 2017.}
	\end{IEEEbiography}

	\vspace{-1.1cm}
	\begin{IEEEbiography}[{\includegraphics[width=1in,height=1.25in,clip,keepaspectratio]{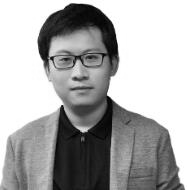}}]{Jianlong Fu} received the Ph.D. degree in pattern recognition and intelligent system from the Institute of Automation, Chinese Academy of Science, in 2015. He is currently a Lead Researcher with the Multimedia Search and Mining Group, Microsoft Research Asia (MSRA). He has authored or coauthored more than 40 papers in journals and conferences, and one book chapter. His current research interests include computer vision, computational photography, vision, and language. He received the Best Paper Award from ACM Multimedia 2018, and has shipped core technologies to a number of Microsoft products, including Windows, Office, Bing Multimedia Search, Azure Media Service, and XiaoIce. He is an Area Chair of ACM Multimedia 2018, ICME 2019. He serves as a Lead organizer and a Guest Editor for the IEEE Transactions on Pattern Analysis and Machine Intelligence Special Issue on Fine-grained Categorization.
	\end{IEEEbiography}
	
	\vspace{-1.1cm}
	\begin{IEEEbiography}[{\includegraphics[width=1in,height=1.25in,clip,keepaspectratio]{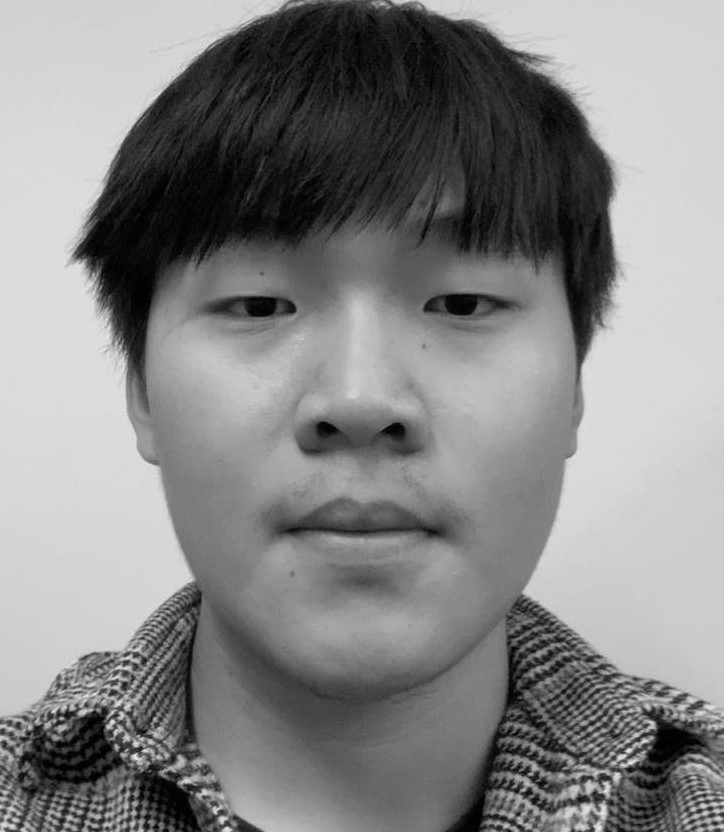}}]{Hao Du} %received the BSc degree from the University of Electronic Science and Technology of China (UESTC), Chengdu, China, in 2019. He 
	is currently pursuing the Ph.D. degree with the City University of Hong Kong (CityU), Hong Kong, China. His research interests include neural architecture design and search and video object tracking.
	\end{IEEEbiography}
	
	\vspace{-0.1cm}
	\begin{IEEEbiography}[{\includegraphics[width=1in,height=1.25in,clip,keepaspectratio]{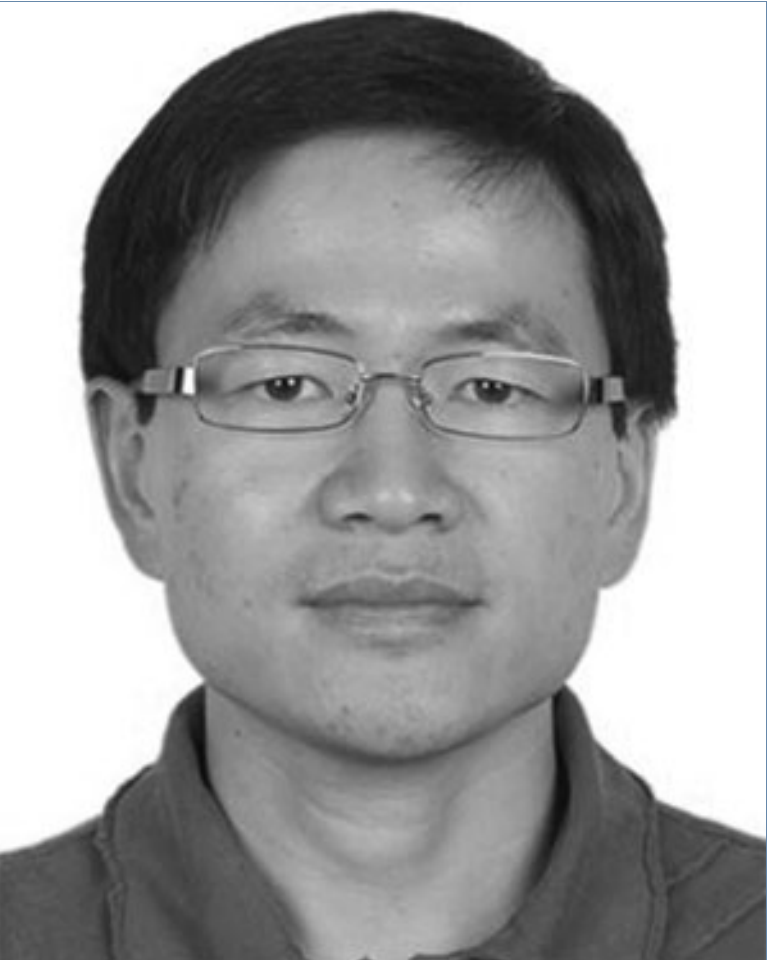}}]{Liang Wang} received the BEng and MEng degrees from Anhui University, in 1997 and 2000, respectively, and the PhD degree from the Institute of Automation, Chinese Academy of Sciences (CASIA), in 2004. From 2004 to 2010, he was a research assistant at Imperial College London, United Kingdom, and Monash University, Australia, a research fellow with the University of Melbourne, Australia, and a lecturer with the University of Bath, United Kingdom, respectively. Currently, he is a full professor of the Hundred Talents Program at the National Lab of Pattern Recognition, CASIA. His major research interests include machine learning, pattern recognition, and computer vision. He has widely published in highly ranked international journals such as the IEEE Transactions on Pattern Analysis and Machine Intelligence and the IEEE Transactions on Image Processing,
		and leading international conferences such as CVPR, ICCV, and ECCV. He is a fellow of the IEEE and the IAPR.
	\end{IEEEbiography}
	
	\vspace{-0.0cm}
	\begin{IEEEbiography}[{\includegraphics[width=1in,height=1.25in,clip,keepaspectratio]{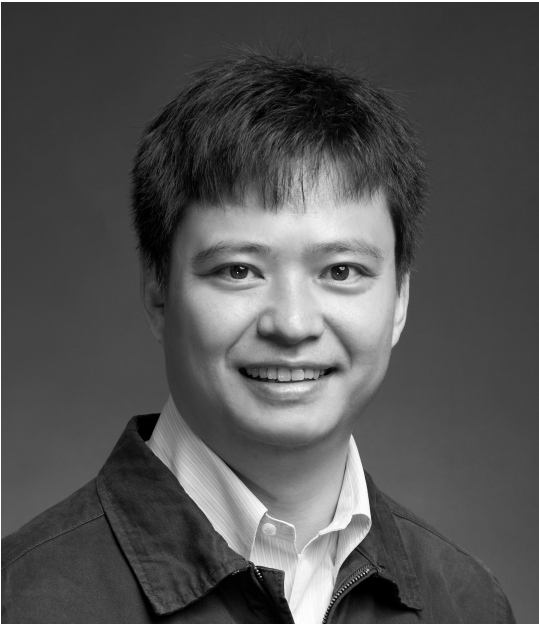}}]{Haibin Ling} received the B.S. and M.S. degrees from Peking University in 1997 and 2000, respectively, and the Ph.D. degree from the University of Maryland, College Park, in 2006. From 2000 to 2001, he was an assistant researcher at Microsoft Research Asia. From 2006 to 2007, he worked as a postdoctoral scientist at the University of California Los Angeles. In 2007, he joined Siemens Corporate Research as a
		research scientist; then, from 2008 to 2019, he worked as a faculty member of the Department of Computer Sciences at Temple University. In fall 2019, he joined Stony Brook University as a SUNY Empire Innovation Professor in the Department of Computer Science. His research interests include computer vision, augmented reality, medical image analysis, and human computer interaction. He received Best Student Paper Award at ACM UIST (2003), NSF CAREER Award (2014), Yahoo Faculty Research Award (2019), and Amazon AWS Machine Learning Research Award (2019). He serves as Associate Editors for several journals including
		IEEE Trans. on Pattern Analysis and Machine Intelligence (PAMI), Pattern Recognition (PR), and Computer Vision and Image Understanding (CVIU), and has served as Area Chairs various times for CVPR and ECCV.
	\end{IEEEbiography}
	
\end{document}